\documentclass[runningheads]{llncs}
\usepackage{graphicx}
\usepackage[utf8]{inputenc}
\usepackage{color,xcolor}
\usepackage{amsmath}
\usepackage{amsfonts}
\usepackage{bbm}
\usepackage{amsmath,amssymb,amsfonts}
\usepackage{comment}
\usepackage{caption}
\usepackage{subcaption}

\usepackage{diagbox}
\usepackage{nameref}

\usepackage{hyperref}
\usepackage{color,xcolor}
\usepackage[ruled,vlined,linesnumbered]{algorithm2e}
\usepackage{algorithm2e}
\usepackage{algorithmic}
\usepackage{multirow}
\usepackage{helvet} % DO NOT CHANGE THIS
\usepackage{courier}  % DO NOT CHANGE THIS
\DeclareMathOperator*{\argmin}{arg\,min}

\newtheorem{assumption}{Assumption}[theorem]
\def\genbox#1#2#3#4#5#6{% #1=0/1, #2=color, #3=shape, #4=raise, #5=width, #6=width/2
	\leavevmode\raise#4bp\hbox to#5bp{\vrule height#5bp depth0bp width0bp
		\pdfliteral{q .5 w \csname #2COLOR\endcsname\space RG
			\csname #3PDF\endcsname{#5}{#6} S Q
			\ifx1#1 q \csname #2COLOR\endcsname\space rg 
			\csname #3PDF\endcsname{#5}{#6} f Q\fi}\hss}}

\title{Attack Transferability Characterization for Adversarially Robust Multi-label Classification}

\begin{document}
%

%\author{anonymous author}
\author {
	% Authors
	Zhuo Yang,\textsuperscript{\rm 1}
	Yufei Han, \textsuperscript{\rm 2}
	Xiangliang Zhang \textsuperscript{\rm 1} \\
}
%% First names are abbreviated in the running head.
%% If there are more than two authors, 'et al.' is used.
%%
\institute{King Abdullah University of Science and Technology, Thuwal, Saudi Arabia
\email{zhuo.yang@kaust.edu.sa, xiangliang.zhang@kaust.edu.sa}\\
% \url{http://www.springer.com/gp/computer-science/lncs} 
\and
CIDRE team, Inria, France\\
\email{yfhan.hust@gmail.com}}
\maketitle              % typeset the header of the contribution
\begin{abstract}
Despite of the pervasive existence of multi-label evasion attack, it is an open yet essential problem to characterize the origin of the adversarial vulnerability of a multi-label learning system and assess its attackability. In this study, we focus on non-targeted evasion attack against multi-label classifiers. The goal of the threat is to cause miss-classification with respect to as many labels as possible, with the same input perturbation. Our work gains in-depth understanding about the multi-label adversarial attack by first characterizing the transferability of the attack based on the functional properties of the multi-label classifier. We unveil how the transferability level of the attack determines the attackability of the classifier via establishing an information-theoretic analysis of the adversarial risk. Furthermore, we propose a transferability-centered attackability assessment, named Soft Attackability Estimator (SAE), to evaluate the intrinsic vulnerability level of the targeted multi-label classifier. This estimator is then integrated as a transferability-tuning regularization term into the multi-label learning paradigm to achieve adversarially robust classification. The experimental study on real-world data echos the theoretical analysis and verify the validity of the transferability-regularized multi-label learning method.

%Intuitively, an attack designed to flip one label is more easy to transfer on other labels if the decision boundaries of these labels align more closely. Our work models this attack transferability into an expected generalization risk bound in the presence of adversary. Specially, the bound is based on conditional mutual information (CMI), which facilitates the exploiting of label correlation on classification risk. Interestingly, we unveils the essential trade-off between the generalization on clean data and perturbed data when encoding label correlation into the learned multi-label classifiers. Based on the theoretical analysis, we also develop a regularization term to inhabit the transfer of attack among labels, thus reducing the attackability of multi-label classifiers. Experiments are organized to verify our theoretical results and the effectiveness of the developed regularizer.

%By intuition, if the decision outputs of different labels are closely correlated, the impact of the attack noise is easy to transfer across the labels. The transferability of the attack thus facilitates the non-targeted threat and harms the utility of the targeted multi-label classifier. To gain in-depth understanding about the adversarial vulnerability, 

\end{abstract}
%总之
%
%
\section{Introduction}\label{sec:introduction}

Adversarial evasion attack against real-world multi-label learning systems can not only harm the system utility, but also facilitate advanced downstreaming cyber meances \cite{SongQi2018ICDM}. For example, hackers embed toxic contents into images while hiding the malicious labels from the detection \cite{Gupta2013www}. Stealthy harassment applications, such as phone call dictation and photo extraction, carefully shape the app function descriptions to evade from the sanitary check of app stores \cite{kevin2020sp,Ristenpart2018chi}. Despite of the threatening impact, it remains an open problem to characterize key factors determining the attackability of a multi-label learning system. Compared to in-depth adversarial vulnerability study of single-label learning problems \cite{YangNeuIPS2020,Fawzi2020MachineLearning,zhang2019icml,Han2020kdd,TuNIPS19}, this is a rarely explored, yet fundamental problem for trustworthy multi-label classification.

%We are inspired to explore this field in this paper.
We focus on the non-targeted evasion attack against multi-label classifiers. In contrast to the single-label learning problem, the goal of the adversarial threat is to kill multiple birds with one stone: it aims at \textit{changing as many label-wise outputs as possible simultaneously, with the same input}. 
%a multi-label classifier is a multiple-output model. One input instance triggers classification outputs with respect to different labels. Therefore, 
% \textcolor{red}{Replace Figure.1 with a simpler example, only focusing on interpreting the impact of label dependency on attack transferability}
\begin{figure}[!t]
	\centerline{\includegraphics[width=0.55\columnwidth,height=2.7cm]{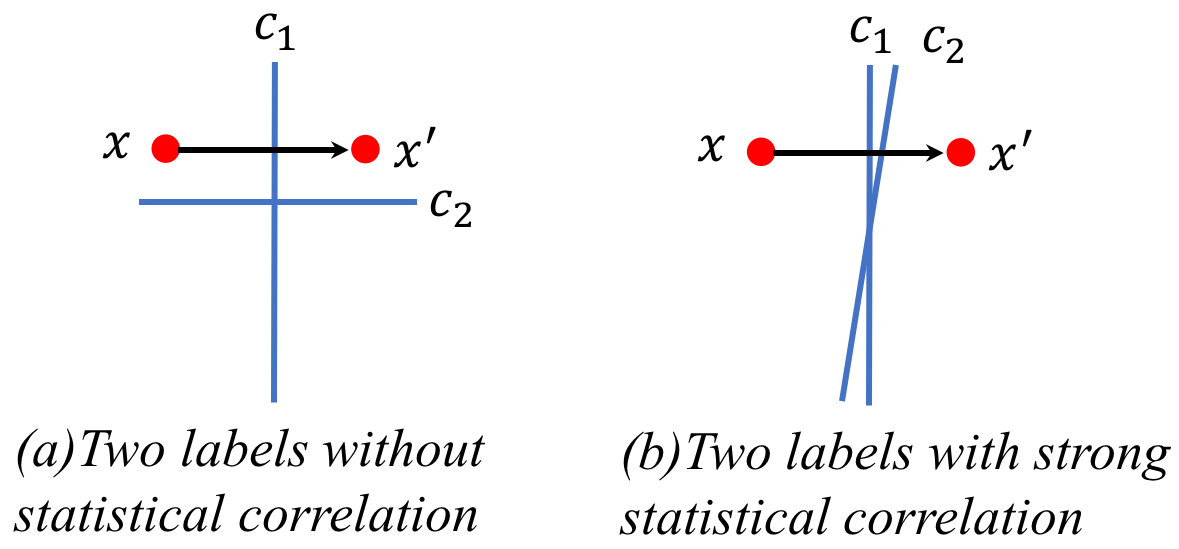}}
	\caption{\textbf{A toy example of multi-label evasion attack.}}
	\label{toy}
\end{figure} 
Fig.\ref{toy} demonstrates a toy example of the threat scenario with two labels $l_{1}$ and $l_{2}$, with decision hyper-planes  $c_{1}$ and $c_{2}$, respectively. Fig.1 (a) assumes no statistical correlation between the two labels. $c_{1}$ and $c_{2}$ are orthogonal therefore. In contrast, the two boundaries are well aligned in Fig.1 (b), implying a strong correlation between $l_{1}$ and $l_{2}$. The injected evasion noise in both scenarios has the same magnitude to change $x$ to be $x'$, indicating the same attack strength. %it has different impacts to the classification system.
%$c_{1}$ and $c_{2}$ are orthogonal in Figure.1(a), indicating bare statistical correlation between the two labels. In contrast, the two boundaries are well aligned in Figure.1(b), implying strong correlation between $l_{1}$ and $l_{2}$. 
As we can see, the evasion attack can flip simultaneously the classifier's output with respect to both labels in Fig.\ref{toy} (b), due to the alignment between the decision boundaries of $l_{1}$ and $l_{2}$. However, in Fig.\ref{toy} (a), the evasion perturbation can only bring impacts to the decision output of $l_{1}$. As shown in the toy example, \textit{whether the attack can transfer across different labels depends on the alignment of the decision hyper-planes, which is determined intrinsically by the correlation between the labels}. On the closely correlated labels, the multi-label classifier tends to produce the consistently same or converse decisions. The adversarial noise that successfully perturbs the decision over one label is likely to cause miss-classification on the other labels. Bearing the goal of the non-targeted attack in mind, the transferability of the attack is closely related to the adversarial vulnerability of the targeted multi-label learning system. With a more transferable attack noise, the multi-label learning system is more attackable.

Given a multi-label learning task, our study aims at gaining in-depth understanding about the theoretical link between the transferability of the evasion attack across different labels and the adversarial vulnerability of the classifier. More specifically, we focus on characterizing the role of attack transferability in determining the adversarial risk of a multi-label  classifier. Furthermore, we pursue a qualitative assessment of attack transferability based on the intrinsic functional properties of the classifier. It is beneficial to not only evaluate the attackability of the classifier, but also design a transferability-suppression regularization term to enhance the adversarial robustness of the classifier. In the community of multi-label learning, it is a well-known fact that capturing the correlation between labels helps to train accurate multi-label classifiers. However, our analysis unveils the other side of the story: encoding the label correlation can also make the classifier vulnerable in the evasion attack scenarios. Our contribution can be summarized as in the followings: 

\begin{itemize}
	\item We unveil the three key factors determining the adversarial risk of a multi-label classifier by establishing an
	information-theoretic upper bound of the adversarial risk. They are i) the conditional mutual information (CMI) between the training data and the learnt classifier \cite{cmi2020}; ii) the transferability level of the attack;
	and  iii) the strength of the evasion attack.
	Theoretical discussions over the first two factors unveil a dilemma: Encoding label correlation in the training data is the key-to-success in accurate adversary-free multi-label classification. However, it also increases the transferability of the attack, which makes the classifier vulnerable (with a higher adversarial risk). %We reveal the dilemma via the theoretical study in Section.\ref{sec:adversarial_bound}. 

	\item  We propose an attackability assessment in Section.\ref{sec:transferability_reg} based on the unveiled link between the attack transferability and the adversarial risk. This attackability assessment is then integrated into the multi-label learning paradigm as a regularization term to suppress the attack transfer and enhance the adversarial robustness of the derived multi-label classifier. 
	%Specially, the regularizer is designed to reduce the empirical attackability measured by a proposed soft attackability measurement method (SAE). SAE works to find the most attackable direction in which an bounded perturbation $\bf{r}$ can flip the most soft labels. Besides, by summing up the damage caused by $\bf{r}$ on each single label,  SAE encodes the attack transferability into the final measurement, which gives the regularizer the ability to inhibit the transfer of attack among labels. 
	
	\item Our empirical study with real-world multi-label learning applications instantiates the theoretical study with both linear and deep multi-label models. The results 
	confirm the trade-off between the utility of the classifier and its adversarial robustness by controlling the attack transferability. 
	%The results verify our theoretical analysis.
\end{itemize}
%transferability-suppression oriented regularization term and propose a regularized training paradigm to enhance the adversarial robustness of multi-label classifiers.

\section{Related work}\label{sec:related}
%\textcolor{blue}{I tried to shrink the discussion of [24] and removed the mention of Eq1, as it is too detailed and too early to mention.  It will be good if we first give some short discussion for related work of  vulnerability analysis about single-label models. For example,}
Bounding adversarial risk has been studied extensively in single-label learning scenarios \cite{Matthias2017nips,AFawzi2016nips,Gilmer2018arxiv,Matthias2017nips,Wang2018AnalyzingTR,Yin2018icml,JKhim2018ft,zhang2019icml,TuNIPS19,Han2020kdd,YangNeuIPS2020,Fawzi2020MachineLearning}. They focus on identifying the upper bound of adversarial noise, which guarantees the stability of the targeted classifier's output, a.k.a. adversarial sphere. Notably, \cite{AFawzi2016nips,Matthias2017nips,Yin2018icml,TuNIPS19} study the association between adversarial robustness and the curvature of the learnt decision boundary. Strengthened further by \cite{Yin2018icml,JKhim2018ft,TuNIPS19}, the expected classification risk under adversarial perturbation can be bounded by the model's Rademacher complexity of the targeted classifier. \cite{zhuo2020} extends the model complexity-dependent analysis to the multi-label learning problems and associates the Rademacher complexity with the nuclear norm of the model parameters. 

Distinguished from single-label learning scenarios, the key-to-success of training an accurate multi-label classifier is to capture the correlation between the labels. More specifically, the alignment between the decision hyper-planes of the correlated labels helps to predict the occurrence of the labels. However, as revealed in \cite{SongQi2018ICDM,zhuo2020}, the evasion attack perturbation can transfer across the correlated labels: \textit{the same input perturbation can affect the decision output of these labels.} It implies that the label correlation can be potentially beneficial to adversaries at the same time. Nevertheless, the relation between the transferability of the input perturbation and the adversarial vulnerability of the victim classifier can not be characterized or measured by the Rademacher-complexity-based analysis conducted on the single-label case and \cite{zhuo2020}. Our work thus focuses on addressing the essential yet open problem from two perspectives. First, we target on establishing a theoretical link between the transferability measurement of the attack noise across multiple labels and the vulnerability of the classifier.
Second, we conduct an information-theoretic  analysis  of  the  adversarial  risk, which is
an attack-strength-independent vulnerability assessment. This assessment can be used to guide proactive model hardening, e.g. robust model training, to improve the adversarial robustness of the classifier.

\section{Vulnerability Assessment of Multi-label Classifiers}\label{sec:adversarial_bound}
%\subsection{Problem Definition}
\noindent \textbf{Notations.} We use $z=(x,y)$ as a multi-label instance, with feature vector ${x} \in {\mathbb{R}^d}$ and label vector ${y} = {\{-1,1\} ^m}$, where $d$ and $m$ denote the feature  dimension  and the number of labels, respectively. Specially, we use $x_i$ and $y^i$ to denote the feature vector and the label vector of instance $z_i$ respectively and use $y_j$ to denote the $j$-th element of label vector $y$. Let $\mathcal{D}$ be the underlying distribution of $z$ and $z^n$ be a data set including $n$ instances. Let $h$ denote the multi-label classifier to learn from the data instances sampled from $\mathcal{D}$. The learning paradigm (possibly randomized) is thus noted as $\mathcal{A}:{z^n} \to {h}$. The probability distribution of the learning paradigm is $\mathcal{P}_{\mathcal{A}}$. The corresponding loss function of $\mathcal{A}$ is $\ell:h \times z \to \mathbb{R}$. %${\bf{x}}_i$ and ${\bf{y}}_i$ are the feature vector and label vector of the i-th instance in $z^{n}$. $h_j$ ($j=1,2,3,...,m$) is the classification output of $h$ for the label $j$ ($j\in{\{1,2,3,...,m\}}$). 
$\|x\|_{p}$ ($p\geq{1}$) denotes the $L_{p}$ norm of a vector $x$. Without loss of generality, we choose $p=2$ hereafter. 

\noindent
\textbf{Attackability of a Multi-label Classifier.} The attackability of $h$ is defined as the expected maximum number of flipped decision outputs by injecting the perturbation ${r}$ to ${x}$ within an attack budget $\varepsilon$:
\begin{equation}\label{eq:attackdf}
\small
\begin{array}{l}
{C^*}(\mathcal{D}) = \mathop \mathbb{E}\limits_{{{z}}\sim\mathcal{D}} \left( {\mathop {{{\max}}}\limits_{T,\left\| {{{{r}}^*}} \right\| \le {\varepsilon}} \,\,\,\sum\limits_{j = 1}^m \mathbbm{1} ({y_j} \ne sgn({h_j}({{x}} + {{{r}}^*})))} \right),\\
{\rm{where}}\,\,\,{{{r}}^*} = \mathop {argmin}\limits_{{r}} \left\| {{r}} \right\|_{p},\,\\
s.t.\,\,{y_j}{{h}_j}({{x}} + {{{r}}^*}) \le 0\,\,(j \in T),\,\,\,{y_j}{{h}_j}({{x}} + {{{r}}^*}) > 0\,(j \notin T).
\end{array}
\end{equation}
$T$ denotes the set of the attacked labels. ${h}_{j}({x}+{r})$ denotes the decision score of the label $j$ of the adversarial input. $\mathbbm{1}(\cdot)$ is the indicator function. It is valued as 1 if the attack flips the decision of the label $j$ and 0 otherwise. With the same input ${x}$ and the same attack strength $\|{r}\|_{p}$, one multi-label classifier $h$ is more vulnerable to the evasion attack than the other $h'$, if $C^{*}_{h}>C^{*}_{h'}$.

%\subsection{Preliminaries}\label{pre}
%\textbf{Conditional Mutual Information (CMI) of the learning paradigm $\mathcal{A}$} \cite{cmi2020}: Let $\tilde Z \in {\mathcal{Z}^{n \times 2}}$ consists of $2n$ instances drawn independently form the distribution $\mathcal{D}$ and organized into a n-by-2 matrix. Each $\tilde Z_{i,j} (i = 1, \cdots ,n; j=1, 2)$ denotes the instance locating at $i$-th row amd $j$-th column of $\mathcal{Z}$.  Let $E \in {\left\{ {1,2} \right\}^n}$ be uniformly sampled with replacement from $\{1,2\}$. Sampling of $E$ is independent from $\mathcal{D}$ and the randomness of the learning paradigm $\mathcal{A}$. Let ${\tilde Z_E} \in {\mathcal{Z}^n}$ by ${({\tilde Z_E})_i} = {\tilde Z_{i,{E_i}}}, (i = 1, \cdots ,n)$. That is, ${\tilde Z_E}$ is the subset of $\tilde Z$ indexed by $E$. The conditional mutual information of $\mathcal{A}$ with respect to $\mathcal{D}$ is
%\begin{equation}\label{eq:cmi}
%\small
%CM{I_{\mathcal{D},\mathcal{A}}}: = I(\mathcal{A}({\tilde Z_E});E|\tilde Z).
%\end{equation}
%$CM{I_{\mathcal{D},\mathcal{A}}}$ measures the amount of the information about the training data instances, which can be unveiled given the classifier $h$ trained using the learning paradigm $\mathcal{A}$. Intuitively, high \textit{CMI} denotes better fit of the classifier $h$ to the training data distribution $\mathcal{D}$. 

\subsection{Information-theoretic Adversarial Risk Bound}\label{method:bound}
Solving Eq.(\ref{eq:attackdf}) directly for a given data instance $z$ reduces to an integer programming problem, as  \cite{zhuo2020} did. Nevertheless, our goal is beyond solely empirically assessing the attackability of $h$ on a given set of  instances. We are interested in \textbf{1)} establishing an upper bound of the expected miss-classification risk of $h$ with the presence of adversary. It is helpful for characterizing the key factors deciding the adversarial risk of $h$; \textbf{2)} understanding the role of the transferability of the input perturbation across different labels in shaping the adversarial threat. 

%We pursue further in-depth understanding about the adversarial vulnerability of $h$. Notably, we 
%over the legal data distribution $\mathcal{D}$ and the attack budget limit $\|\bf{r}\|_{p}\leq{\varepsilon}$
%empirically evaluating $C^{*}(h)$ in Eq.\ref{eq:attackdf} for a given set of data instance $z^{n}$, 
%Following Eq.(\ref{eq:attackdf}) as the measurement of evasion attackability of a multi-label classiffier, we further cast the attackability analyzing into the generalization theory framework.
%That is, we pursue the insights of potential misclassification risk faced by the learned classifier when it encounters new data and designed attack perturbation. Specially, we are interested to the transferability of attack among labels and the role of label dependency in the attackability measurement.
%on attack transferability and model  generalization ability on clean unseen data.  

For a multi-label classifier $h$, $n$ legal instances $z^{n}=\{z_{i}\}$ ($i$=$1,2,...,n$, $z_{i}=({x}_{i},{y}^i)\sim{\mathcal{D}}$) and the attack budget $\varepsilon$, we can estimate the expected adversarial risk of $h$ by evaluating the worst-case classification risk over the neighborhood $N({{z}_{i}}) = \left\{ {({{x'}_{i}},{{y}^i})\left| {{{\left\| {{{x'}_{i}} - {{x}_{i}}} \right\|}_p} \le \varepsilon } \right.} \right\}$. The expected and empirical adversarial risk $R_{\mathcal{D}}(h,\varepsilon)$ and $R^{emp}_{\mathcal{D}}(h,\varepsilon)$ give: 
% \textcolor{blue}{Should we use here $y^i$, not $y_i$, because in Eq.1 $y_j$ is the $j$-th label of $x$. We can use $y^i$ to denote the label vector of instance $x_i$. And what is $\mathcal{D}^{n}$?}
\begin{equation}\label{eq:emrisk}
\small
\begin{split}
{R_{\mathcal{D}}}(h,\varepsilon) &= {E_{\mathcal{A},{z}^{n}\sim{\mathcal{D}^{n}}}} {[E_{{z}\sim{\mathcal{D}}}[\mathop {\max }\limits_{{({x}',{y})\in{N({z})}}} \ell(h({{{x}'}}),{{y}})]]} ,\,\,\,h=\mathcal{A}(z^{n}),\\
{R^{emp}_{\mathcal{D}}}(h,\varepsilon) &= E_{\mathcal{A},{z}^{n}\sim{\mathcal{D}^{n}}}[\frac{1}{n}\sum\limits_{i = 1}^n {[\mathop {\max }\limits_{{{({x}'_{i},{y}^i)\in{N({z}_{i})}}}} \ell(h({{{x}'}_i}),{{{y}}^i})]]},\,\,\,h=\mathcal{A}(z^{n}).\\
\end{split}
\end{equation}

The expectation in Eq.(\ref{eq:emrisk}) is taken with respect to the joint distribution ${\mathcal{D}^{ \otimes n}} \otimes {\mathcal{P}_{\mathcal{A}}}$ and ${\mathcal{D}^{ n}}$ denotes the data distribution with $n$ instances. The expected adversarial risk $R_{\mathcal{D}}(h,\varepsilon)$ reflects the vulnerability level of the trained classifier $h$. Intuitively, a higher $R_{\mathcal{D}}(h,\varepsilon)$ indicates that the classifier $h$ trained with the learning paradigm $\mathcal{A}$ is easier to attack (more attackable). %Thus $h=\mathcal{A}(\mathcal{D}^{n})$ is more attackable.for each ${z}_{i} = ({x}_i,{y}_i)$
${R^{emp}_{\mathcal{D}}}(h,\varepsilon)$ is the empirical evaluation of the attackability level. By definition, if $\mathcal{A}$ is deterministic and the binary 0-1 loss is adopted, $\sum_{i=1}^{n} C^{*}_{h}({ z}_{i})$ gives ${R^{emp}_{\mathcal{D}}}(h,\varepsilon)$. 
%Furthermore, the difference between Eq.(\ref{eq:trisk}) and Eq.(\ref{eq:erisk}) is called adversarial generalization risk, i.e $gen(A,\varepsilon )$. With $gen(A,\varepsilon )$, the expected adversarial true loss can be decomposed as $ {R_\mathcal{D}}(A,\varepsilon ) = {R_{Z^n}}(A,\varepsilon ) + gen(A,\varepsilon )$, where the first term reflects how well the output hypothesis fits the dataset and the second term reflects how well the output hypothesis genneralizes. To minimize ${R_\mathcal{D}}(A,\varepsilon )$, we need the both terms ${R_{Z^n}}(A,\varepsilon )$ and $gen(A,\varepsilon )$ to be small.

Theorem.\ref{bound:general} establishes the upper bound of the adversarial risk ${R_\mathcal{D}}(h,\varepsilon )$ based on the conditional mutual information $CM{I_{\mathcal{D},\mathcal{A}}}$ between the legal data and the learning paradigm. Without loss of 
generality, the hinge loss is adopted to compute the miss-classification risk of each ${z}$, i.e., $\ell({h},{{z}=({x},{y})})= \sum\nolimits_{j = 1}^m{\max \{ 0,1 }$ ${- {y_j}{{h}_j}({{x}})\} } $. We consider one of the most popularly used structures of multi-label classifiers, i.e., $h({{x}}) = {\bf{W}} Rep({{x}})$, where ${\bf{W}} \in R^{m*d'}$ is the weight of a linear layer and $Rep({{x}})\in{R^{d'}}$ is a $d'$-dimensional representation vector of  ${x}\in{R^{d}}$, e.g., from a non-linear network architecture.
%$Rep({\bf{x}})$ returns $\bf{x}$ to make $H({\bf{x}}) $ a linear multi-label classifier. 
In Theorem.\ref{bound:general}, we assume a linear hypothesis $h$, i.e., $Rep({{x}}) = x$ for the convenience of analysis. The conclusion holds for more advanced architectures, such as feed-forward neural networks. % with $h({{x}}) = {\bf{W}} \cdot Rep({{x}})$. 

%We further show that the conclusion holds when $Rep({\bf{x}})$ is a nonlinear feature mapping.
%\textcolor{red}{incude together?}. 
\begin{theorem}\label{bound:general}
	%\textbf{[Upper bound of attackability for linear hypotheses].}
	%Let $A:{\mathcal{Z}^n} \to \mathcal{H}$ and $\mathcal{D}$ be the distribution on $\mathcal{Z}$. Let $l:\mathcal{H} \times \mathcal{Z} \to [0, + \infty )$ be the hinge loss. 
	Let $h = {\bf{W}}x$ be a linear multi-label classifier. We further denote $\mathcal{D}=(\mathcal{D}_1, \cdots, \mathcal{D}_m)$ and ${\bf{W}} = ({{\bf{w}}_1}, \cdots ,{{\bf{w}}_m})$, where $\mathcal{D}_j$ is the data distribution w.r.t. each label $j$ and ${\bf{w}}_j$ is the weight vector of the classifier of   label $j$. 
	\begin{equation}\label{eq:tbound}
	\small
	\begin{array}{ll}
	{R_\mathcal{D}}(h,\varepsilon )  \le & {R^{emp}_{{\mathcal{D}}}}(h,\varepsilon ) + \\
	&  {\left(\frac{2}{n}CM{I_{\mathcal{D},A}}  \mathop \mathbb{E}\limits_{{z}=({{x}},{{y}}) \sim \mathcal{D}} \left[ {\mathop {\sup }\limits_{{\bf{W}} \in \mathcal{W}_A} {{\left( {l({\bf{W}},z) + {C_{{\bf{W}},z}}  \varepsilon } \right)}^2}} \right]\right)}^{1/2} 
	\end{array},
	\end{equation}
	where $\mathcal{W}_\mathcal{A}$ is the set including all possible weight vectors learned by $\mathcal{A}$ using the data set $z^n$ sampled from $\mathcal{D}^{n}$. ${C_{{\bf{W}},{z}}} = \mathop {\max }\limits_{\{ {b_1}, \cdots ,{b_m}\} } {\left\| {\sum\nolimits_{j = 1}^m {{b_j}{y_j}{{\bf{w}}_j}} } \right\|_2},{b_j} = \{ 0,1\} $.
%\bf{W}} \leftarrow \mathcal{A}({z^n})
% \textcolor{blue}{Should ${z}=({{x}},{{y}}) \leftarrow \mathcal{D}$ in Eq.3$\&$4 be ${z}\sim{\mathcal{D}}$ or ${z}^n\sim{\mathcal{D}^n}$ ? like in Eq2?}	
	The empirical adversarial risk ${R_{{Z^n}}}(A,\varepsilon )$ has the upper bound:
	\begin{equation}\label{eq:ebound}
	\small
	{R^{emp}_{{\mathcal{D}}}}(h,\varepsilon ) \le {R^{emp}_{{\mathcal{D}}}}(h,0) + \mathop \mathbb{E}\limits_{{z^n} \sim {\mathcal{D}^n},\mathcal{A}} \left[ {\mathop {\sup }\limits_{\bf{W}\in{\mathcal{W}_\mathcal{A}}} \mathop \mathbb{E}\limits_{z \in {z^n}} \left( {{C_{{\bf{W}},z}}  \varepsilon } \right)} \right],
	\end{equation}
	where ${R^{emp}_{{\mathcal{D}}}}(h,0)$ denotes the empirical and adversarial-free classification risk.
	
%	For a more intuition of the influence of the label dependency on measured attackability, 
	\sloppy We further provide the upper bound of $CM{I_{\mathcal{D},\mathcal{A}}}$ as:
% 	\textcolor{red}{In Eq.6, what does k denote?}
	\begin{equation}\label{eq:cmibound}
	\small
	%CM{I_{{\cal D},A}} \le ent({{\bf{w}}_1}, \cdots ,{{\bf{w}}_m}) + ent({\cal D}_1^{n \times 2}, \cdots ,{\cal D}_m^{n \times 2})
	CM{I_{{\cal \mathcal{D}},\mathcal{A}}} \le ent({{\bf{w}}_1}, \cdots ,{{\bf{w}}_m}) + ent({\cal \mathcal{D}}_1, \cdots ,{\cal \mathcal{D}}_m)
	\end{equation}
	where $ent(\cdot)$ denotes the entropy of the concerned random variables. 
	
%${{\cal D}^{n \times 2}} = ({\cal D}_1^{n \times 2}, \cdots ,{\cal D}_m^{n \times 2})$ denotes the distribution of data set with $n \times 2$ instances and
% 	Note that term $ \sum\nolimits_{i = 1}^m {ent(\mathcal{D}_k^{n \times 2}\left| {\mathcal{D}_{k - 1}^{n \times 2}, \cdots ,\mathcal{D}_1^{n \times 2}} \right.)}$ can be used to measure the label dependency among involved labels.
\end{theorem}

% \textbf{Proof of Eq.(\ref{eq:cmibound}):} Here we use $H$ to denote the entropy.
% \begin{equation}
% \begin{array}{l}
% CM{I_{\mathcal{D},A}}\\
%  = I(A;S,\bar Z) - I(A;\bar Z)\\
%  = H(A) + H(S,\bar Z) - H(A,S,\bar Z) - H(A) - H(\bar Z) + H(A,\bar Z)\\
%  = H(A,\bar Z) + H(S|\bar Z) - H(S) - H(A,\bar Z|S)\,\,\,\,:S\,\,is\,independent\,to\,Z\\
%  = H(A,\bar Z) - H(A,\bar Z|S)\\
%  \le H(A,\bar Z)\\
%  \le H(A) + H(\bar Z)\\
%  = H({\bf{W}}) + H(\bar Z)\\
%  = \sum\limits_{i = 1}^m {H({{\bf{w}}_k}\left| {{{\bf{w}}_{k - 1,}} \cdots ,{{\bf{w}}_1}} \right.)}  + \sum\limits_{i = 1}^m {H(\mathcal{D}_k^{n \times 2}\left| {\mathcal{D}_{k - 1}^{n \times 2} \cdots ,\mathcal{D}_1^{n \times 2}} \right.)} 
% \end{array}
% \end{equation}

%The proofs of Eq.(\ref{eq:tbound} - \ref{eq:cmibound}) are presented in supplementary document. 

\noindent
\textbf{Key Factors of Attackability.}
The three key factors determining the adversarial risk (thus the attackability level) of the targeted multi-label classifier are: 1) $CM{I_{\mathcal{D},\mathcal{A}}}$; 2) $\mathop \mathbb{E}\limits_{z} {C_{{\bf{W}},z}}$ (
$\mathop \mathbb{E}\limits_{z \leftarrow \mathcal{D}} {C_{{\bf{W}},z}}$ in Eq.(\ref{eq:tbound}) and
$\mathop \mathbb{E}\limits_{z \in {z^n}} {C_{{\bf{W}},z}}$ in Eq.(\ref{eq:ebound})); 
% ${C_{{\bf{W}},Z}}$ 
and 3) the attack budget $\varepsilon$. % are the three key factors determining the adversarial risk (thus the attackability level) of the targeted multi-label classifier.
% \textcolor{blue}{${z}\sim{\mathcal{D}}$ rather than $z \leftarrow \mathcal{D}$?}

The last factor of the  attack budget $\varepsilon$ is easy to understand. The targeted classifier is intuitively attackable if the adversary has more attack budget. The larger $\varepsilon$ is, the stronger the attack becomes and the adversarial risk rises accordingly.
We then analyze the first factor $CM{I_{\mathcal{D},\mathcal{A}}}$.
For a multi-label classifier $h$ accurately capturing the label correlation in the training data, the output from $h_{j}$ and $h_{k}$ are closely aligned w.r.t. the positively or negatively correlated labels $j$ and $k$. Specifically, in the linear case, the alignment between $h_{j}$ and $h_{k}$ can be presented by $ s(h_{j},h_{k})$=$ \max{\{cos\left\langle{{{\bf{w}}_j},{{\bf{w}}_k}} \right\rangle, cos\left\langle {{-{\bf{w}}_j},{{\bf{w}}_k}} \right\rangle\}}$, where $cos\left\langle *,*\right\rangle$ denotes the cosine similarity. As shown in Eq.(\ref{eq:cmibound}), the alignment of the decision hyper-planes of the correlated labels reduce the uncertainty of $\bf{W} = \mathcal{A}(\mathcal{D})$. Correspondingly, the conditional mutual information $CM{I_{{\cal \mathcal{D}},\mathcal{A}}}$ decreases if the label correlation is strong and the classifier perfectly encodes the correlation into the alignment of the label-wise decision hyper-planes. According to Eq.(\ref{eq:tbound}), it is consistent with the well recognized fact of adversary-free multi-label learning: encoding the label correlation in the classifier helps to achieve an accurate adversary-free multi-label classification.  %a lower $CM{I_{\mathcal{D},\mathcal{A}}}$ indicates better model fit to the training data and better adversary-free accuracy \cite{cmi2020,mi}
\begin{lemma}\label{lemma:cwz}
%Let $\bf{z} = (x,{y}) $ the independent and identically distributed training instance sampled from the data distribution $\mathcal{D}$. $y_{i}=+/-1$ ($i=1,2,3,...,m$) denotes the sign of the label $i$ of $z$. 
$\mathop \mathbb{E}\limits_{z}C_{{\bf{W}},z}$ reaches the maximum value, if for each pair of labels $j$ and $k$, $\mathop \mathbb{E}\limits_{z} \{cos\left\langle {{y_{j}{\bf{w}}_j},{y_{k}{\bf{w}}_k}} \right\rangle\} = 1$.  
\end{lemma}

The second factor $\mathop \mathbb{E}\limits_{z}C_{{\bf{W}},z}$ \textit{measures the transferability of the attack noise and demonstrates the impact of the transferability level on the attackability of the classifier}. With Lemma.\ref{lemma:cwz}, we make the following analysis.
\textbf{First}, for two labels $j$ and $k$ with strong positive or negative correlation in the training data, a large value of $\mathop \mathbb{E}\limits_{z} \{cos\left\langle {{y_{j}{\bf{w}}_j},{y_{k}{\bf{w}}_k}} \right\rangle\}$ indicates a high intensity of $ s(h_{j},h_{k}) = \max{\{cos\left\langle {{{\bf{w}}_j},{{\bf{w}}_k}} \right\rangle,}$ $ {cos\left\langle {{-{\bf{w}}_j},{{\bf{w}}_k}} \right\rangle\}}$. It represents that the decision hyper-planes ${\bf{w}}_{j}$ and ${\bf{w}}_{k}$ of the classifier $h$ are consistently aligned. Therefore, with the same attack strength encoded by $\|r\|_{2}\leq{\varepsilon}$, the adversarial sample $x' = x+r$ tends to cause miss-classification on both $h_{j}(x')$ and $h_{k}(x')$. Therefore, the attack perturbation's impact is easy to transfer between the correlated labels. Otherwise, $\mathop \mathbb{E}\limits_{z} \{cos\left\langle {{y_{j}{\bf{w}}_j},{y_{k}{\bf{w}}_k}} \right\rangle\} = 0$ indicates an orthogonal pair of ${\bf{w}}_{j}$ and ${\bf{w}}_{k}$. The adversarial perturbation $r$ may cause miss-classification on one of the labels, but induce little bias to the decision output of the other. The attack can not be transferred between the labels. Therefore, a higher / lower $\mathop \mathbb{E}\limits_{z}C_{{\bf{W}},z}$ denotes higher / lower transferability of the attack perturbation. \textbf{Second}, according to Eq.(\ref{eq:tbound}) and Eq.(\ref{eq:ebound}), with an increasingly higher $\mathop \mathbb{E}\limits_{z} \{cos\left\langle {{y_{j}{\bf{w}}_j},{y_{k}{\bf{w}}_k}} \right\rangle\}$, the adversarial risk of the targeted classifier $h$ rises given a fixed attack budget $\varepsilon$. In summary, the alignment between the classifier's decision hyper-planes of different labels captures the label correlation. The alignment facilitates the attack to transfer across the labels. A multi-label classifier is more attackable if the attack is more transferable across the labels, as the attack can impact the decision of more labels at the same time.

\begin{remark}\label{ob:trade}
	\textbf{Trade-off between the generalization capability of the classifier on clean data and its adversarial robustness.}

Capturing the label correlation in the learnt multi-label classifier can be a double-edged sword. \textbf{On one hand}, encouraging alignment between the decision hyper-planes of the correlated labels reduces $CM{I_{{\cal \mathcal{D}},\mathcal{A}}}$ under the adversary-free scenario ($\varepsilon=0$ in Eq.(\ref{eq:tbound})), thus reduces the expected miss-classification risk. \textbf{On the other hand}, the alignment between the decision hyper-planes increases the transferability of the attack, which makes the classifier more vulnerable. \textit{Controlling the alignment between the decision outputs of different labels can tune the trade-off between the utility and the adversarial robustness of the classifier}.

\end{remark}

%The discussion and remark above are derived with the setting of linear hypothesis, and they still hold for deep nonlinear models, i.e $Rep(\bf{x})$ is a deep representation learning network. In this case, we can define the sensitivity of $Rep(\bf{x})$ as
%\begin{equation}\label{eq:sen}
%{\Delta _{Rep}} = \mathop {\sup }\limits_{{\bf{x}},{\bf{x'}} \in \mathcal{X},{{\left\| {{\bf{x}} - {\bf{x'}}} \right\|}_2} \le \varepsilon } {\left\| {Rep({\bf{x}}) - Rep({\bf{x'}})} \right\|_2}.
%\end{equation}
%We then replace the attack budget $\varepsilon$ in Eq.(\ref{eq:tbound}) and Eq.(\ref{eq:ebound}) by ${\Delta _{Rep}}$, and the conclusions keep holding.

\section{Transferrability Regularization for Adversarially Robust Multi-label Classification}\label{sec:transferability_reg}

Following the above  discussion, an intuitive solution to achieve adversarially robust multi-label classification is to regularize $\mathop \mathbb{E}\limits_{z \in {z^n}} {C_{{\bf{W}},z}}$ empirically, while minimizing the multi-label classification loss over the training data set $z^{n}$. We denote this training paradigm as \textbf{ARM-Primal}: 
\begin{equation}\label{eq:arm_primal}
\small
%\begin{split}
    h^{*} = \underset{h}{\argmin}\,\,\frac{1}{n}\ell(h,z_{i}) + \frac{\lambda}{n}\sum_{i=1}^{n}C_{{\bf{W}},z_{i}}\\
   %&C_{\bf{W},z_{i}} = \underset{\{b_{1},\cdots,b_{m}\},b_{i}=\{0,1\}}{max}\|\sum_{i=1}^{m}b_{i}y_{i}\bf{w}_{i}\|\\
%\end{split}
\end{equation}
where $\lambda$ is the penalty parameter, and  $C_{{\bf{W}},z_{i}}$ is given as in Theorem.\ref{bound:general}. %${\bf{W}}$ denotes the linear layer of the classifier $h(x)={\bf{W}} Rep(x)$. %$y_{i,j}$ denotes the $j$-th label of the data instance $z_{i} = (x_{i},y_{i})$.= ({{\bf{w}}_1}, \cdots ,{{\bf{w}}_m})
%Concretely, ARM-primal is conducted by training with objective function $l = {l_c} + \lambda \mathop \mathbb{E}\limits_{Z \in {Z^n}} {C_{W,Z}}$, where $l_c$ is the classification loss and $\lambda$ is the hyperparameter to control the strength of regularization. 
%\textcolor{red}{$l = {l_c} + \lambda \mathop \mathbb{E}\limits_{Z \in {Z^n}} {C_{{\bf{W}},Z}}$ how would you conduct ARM-primal based robust learning? Could you provide a formulation of ARM-primal? If you want to regularize the term $C_{W,z}$, how would you define the regularization in a commutable way? }%Note that for a nonlinear model, we can linearize it to obtain ${\bf{w}}_k$ in $\mathop \mathbb{E}\limits_{Z \in {Z^n}} {C_{{\bf{W}},Z}}$. 
As discussed in Section.\ref{sec:adversarial_bound}, the magnitude of ${C_{{\bf{W}},z_{i}}}$ in Eq.(\ref{eq:arm_primal}) reflects the alignment between the classifier's parameters $\{{\bf{w}}_{1},...,{\bf{w}}_{m}\}$. Penalizing large ${C_{{\bf{W}},z_{i}}}$ thus reduces the transferability of the input attack manipulation among different labels, which makes the learnt classifier $h$ more robust against the adversarial perturbation. 
However, \textit{ARM-Primal} only considers the alignment between the parameters of the linear layer ${\bf{w}}_{j}$ ($j$=$1,...,m$). This setting limits the flexibility of the regularization scheme from two perspectives. First, whether $h$ is attackable given a bounded attack budget also depends on the magnitude of the classification margin of the input instance \cite{margin,margin2}. Second, the regularization is only enforced over the linear layer's parameters of $h$. However, it is possible that the other layers could be relevant with the transferability of the attack noise. Adjusting the parameters of these layers can also help to control the attackability.

As an echo, we address accordingly the limits of \textit{ARM-Primal}: \textbf{First}, a soft attackability estimator (\textbf{SAE}) for the targeted multi-label classifier $h$ is proposed to relax the NP-hard attackability assessment in Eq.(\ref{eq:attackdf}). We show that the proposed \textit{SAE} assesses quantitatively the transferabiltiy level of the input attack noise by considering both the alignment of the decision boundaries and the classification margin of the input data instance. The attackability of the classifier is unveiled to be proportional to the transferability of the attack. \textbf{Second}, \textit{SAE} is then introduced as a regularization term to achieve a tunable trade-off between transferability control and classification accuracy of the targeted classifier $h$. It thus reaches a customized balance between adversarial attackability and utility of $h$ for multi-label learning practices. %The \textit{SAE} is designed to measuring the overall impact of all the parameters in $h$ on the transferability of the attack. 
% this sentence is commented, because it has been previously said: SAE assesses quantitatively the transferabiltiy level of the input attack noise by considering both the alignment of the decision boundaries and the classification margin of the input data instance.

% \textcolor{red}{I can't get any idea why ARM-Primal has a limitation / unaddressed issue from what you describe. I think you should give more clear explanation with mathematical formulas here.}
%Let $h$ be the classifier without the last sigmoid layer, i.e $h = {\bf{Wx}}$ for linear classifier. 
\subsection{Soft Attackability Estimator (SAE)}
%Eq.(\ref{eq:attackdf}) can be viewed as a hard empirical attackability measurement method and it mainly serves as a metric to evaluate the adversarial robustness of multi-label classifiers. We then propose a soft empirical attackability measurement method (SAE), which is defined directly over the model parameters and thus used mainly as an optimization object. 
We first introduce the concept of \textit{SAE} with the single-label classification setting and then extend it to the multi-label case.
Suppose $h$ is a   binary classifier   and instance ${x}$ %as a correctly classified instance. ${x}$ 
is predicted as positive if $h{\rm{(}}{{x}}{\rm{) > 0}}$ and vice versa. Let the adversarial perturbation be decomposed as ${r} = c{\tilde{r}}$, where $c=\|{r}\|_{p}$ and $\|{\tilde{r}}\|_{p}=1$, i.e., ${\tilde{r}}$ shows the direction   of the attack noise and $c$ indicates the strength of the attack along this direction.
For the perturbed input ${{x'}} = {{x}} + c{{\tilde{r}}}$, %denote the adversarially perturbed input to the classifier.  T
the first-order approximation of $h({x}')$ is given as: 
\begin{equation}\label{eq:linearize}
\small
%\begin{split}
h({{x}} + c{{\tilde{r}}}) = h({{x}}) + c{{\tilde{r}}}^{T}\nabla h({{x}}),\,\,\,s.t.\,\,\,\|{\tilde{r}}\|_{p}=1,\,\,\,c\geq{0}\\
%\end{split}
\end{equation}
where $\nabla h({{x}})$ denotes the gradient of $h$ to ${x}$.
%Note that we can linearize $h$ when it is a non-linear function. 
%Now we measure the attackability of $h$ over perturbation $\bf{r}$. 
% The magnitude of $\|\bf{r}\|$ required to flip the classifier's output is then used as the measurement of the attackability of the classifier. Higher $\|\|$
% Denote the number as $c$, then we have
To deliver the attack successfully, the magnitude of the attack noise follows: 
%\begin{equation}\label{eq:c}
%h({\bf{x}}{\rm{ + }}c{\bf{r}}) = 0.
%\end{equation} 
%Combine Eq.(\ref{eq:c}) with Eq.(\ref{eq:linearize}), we get
\begin{equation}\label{eq:cc}
\small
c \geq \frac{{ - h({{x}})}}{{{{\tilde{r}}}^{T}\nabla h({{x}})}}.
\end{equation} 
The attackability of $h$ on ${x}$ along the direction of ${\tilde{r}}$ is proportional to $\frac{1}{c}$. The smaller $c$ is, the more attackable the classifier $h$ becomes. %With the single-label setting, we can assume that $\bf{r}$ denotes an attackable direction, which means that the attack must flip the label in the direction of $\bf{r}$ with unlimited attack strength. This assumption keeps $\frac{1}{c} \ge 0$ holding always. 
% bigger $c$ indicates that $h$ is less attackable by $\bf{r}$. 
% While using $c$ to denote the attackability of $H$ by $\bf{r}$ has a problem that $c$ can take negative values. A negative $c$ indicates that $r$ is actually making $h$ more robust. To address this problem, we add a sign function to Eq.(\ref{eq:cc}) and keep $c \ge 0$. We then use $1/c$ to measure the attackability for a positive correlation. 
%We therefore define a continuous scoring scheme evaluating the \textbf{$h$'s attackability level at $\bf{x}$ along the direction of $\bf{\tilde{r}}$} as:
%\begin{equation}\label{eq:attackr}
%\small
%{A_{h({\bf{x}}),{\bf{\tilde{r}}}}} = \frac{{ - {\bf{\tilde{r}^{T}}}\nabla h({\bf{x}})}}{{h({\bf{x}})}}. 
%\end{equation}
 %${A_{h({\bf{x}}),{\bf{r}}}}$ evaluates how many labels (not necessary an integer) an perturbation with unit attack strength can flip in the direction of $\bf{r}$ and is thus called a soft attackability measurement. 
%We finally define the soft attackability of $h$ on instance $\bf{x}$ as:
%\begin{equation}\label{eq:sasingle}
%{\begin{array}{*{20}{l}}
% 	{{\phi _{h,{\bf{x}}}}  = \mathop {\max }\limits_r {A_{H(x),r}},}\\
% 	{s.t.\,\,\,{{\left\| r \right\|}_p} = 1}
% 	\end{array}}
%\end{equation}
%${\phi _{h,{\bf{x}}}}$ evaluates the max number of labels $\bf{r}$ can flip with the constraint that $\bf{r}$ has unit $L_p$ norm. 

Extending the notions to the multi-label setting, we define %the attackability measurement of 
the multi-label classifier \textbf{$h$'s attackability at ${x}$ along the direction of ${\tilde{r}}$}:
\begin{equation}\label{eq:mattackr}
\small
{A_{h({{x}}),{{\tilde{r}}}}} = \sum\limits_{j = 1}^m {\max \{ \frac{{ - {{\tilde{r}}}^{T}\nabla {h_j}({{x}})}}{{{h_j}({{x}})}},0\} } . 
\end{equation}
%where the multi-label classifier as $h = ({h_1},{h_2}, \cdots, {h_m}) $ and $m$ is the number of labels. 
Note that in the multi-label setting, the adversarial perturbation ${\tilde{r}}$ may cause miss-classification of ${x}$ for some labels, while enhancing the correct classification confidence for other labels, i.e.,   %$A_{h_j({x}),{\tilde{r}}}$ 
$\frac{{ - {{\tilde{r}}}^{T}\nabla {h_j}({{x}})}}{{{h_j}({{x}})}}$
can be negative for the labels with enhanced correct classification confidences. We set the corresponding attackability level to 0, as the attack perturbation fails to cause miss-classification. 

The intensity of ${A_{h({{x}}),{{\tilde{r}}}}}$ is proportional to the number of the labels whose   decision outputs are flipped by the perturbation ${\tilde{r}}$. Compared to the hard-count based attackability measurement $C^{*}_{h}$ in Eq.(\ref{eq:attackdf}), ${A_{h({{x}}),{{\tilde{r}}}}}$ is a soft score quantifying the impact of the attack perturbation over the outputs of the classifier. It is therefore regarded as a \textit{soft attackability estimator}. 

%evaluates how many labels (soft count) an perturbation with unit attack strength can flip in the direction of $\bf{r}$, and \textbf{the summation operation in Eq.(\ref{eq:mattackr}) expresses explicitely the transfer of attack}. 

\noindent
\textbf{Transferrability defines attackability.} 
For simplicity, we denote $\frac{{ - \nabla {h_j}({{x}})}}{{{h_j}({{x}})}}$ as ${{\bf{a}}_j}$, and ${A_{h({{x}}),{{\tilde{r}}}}}$ can be further described as
\begin{equation}\label{eq:mattackc_v1}
\small
\begin{array}{*{20}{l}}
{{A_{h({{x}}),{{\tilde{r}}}}} = {{\tilde{r}}}^{T}\sum\limits_{j \in S,S = \left\{ {j;{\rm{sgn}}( - {y_{j}}{{\tilde{r}}}^{T}\nabla {h_j}({{x}})) > 0} \right\}} {{{\bf{a}}_j}} }\\
%{ = {{\left\| {\bf{\tilde{r}}} \right\|}_2}*{{\left\| {\sum\limits_{i \in S} {{{\bf{a}}_i}} } \right\|}_2}*\cos \left. {\left\langle {{\bf{\tilde{r}}},\sum\limits_{i \in S} {{{\bf{a}}_i}} } \right.} \right\rangle }\\
{ = {{\left\| {{\tilde{r}}} \right\|}_2}\sqrt {\sum\limits_{j \in S} {\left\| {{{\bf{a}}_j}} \right\|_2^2 + 2\sum\limits_{j < k;j,k \in S} {{{\left\| {{{\bf{a}}_j}} \right\|}_2}{{\left\| {{{\bf{a}}_k}} \right\|}_2}\cos \left\langle {{{\bf{a}}_j}, {{{\bf{a}}_k}}} \right\rangle } } } \cos \left. {\left\langle {{{\tilde{r}}},\sum\limits_{j \in S} {{{\bf{a}}_j}} } \right.} \right\rangle }
\end{array}
\end{equation}
As shown in Eq.(\ref{eq:mattackc_v1}), the tranferability of the attack noise ${\tilde{r}}$ is measured by the cosine similarity between ${{\bf{a}}_j}$ and ${{\bf{a}}_k}$. Each ${{\bf{a}}_j}$ aligns with the principal eigen-vector of the Fisher Information Matrix (FIM) of $h_{j}$ at the input instance ${x}$ \cite{zhao2019aaai}. It depicts the local geometrical profile of the decision boundaries of different labels near ${x}$. A larger cosine similarity between ${{\bf{a}}_j}$ and ${{\bf{a}}_k}$ indicates a stronger alignment of the decision boundaries of label $j$ and $k$ within the neighborhood of ${x}$. The attack noise ${\tilde{r}}$ thus causes   closer magnitude of perturbation over $h_{j}({x})$ and $h_{k}({x})$ according to Eq.(\ref{eq:mattackr}). It confirms the association between the transferability and the attackability, as unveiled by Eq.(\ref{eq:tbound}) and Eq.(\ref{eq:ebound}). Besides, the magnitude of the gradient ${\bf{a}_{k}}=\nabla{h_{k}({x})}$ also shapes the attackability level. A larger norm $\|\nabla{h_{k}({x})}\|_{2}$ indicates a less stable classification output within the $L_{p}$-ball centered at ${x}$, i.e., a higher attackability level of the classifier. Integrating both factors, $A_{h({x}),{\tilde{r}}}$ 
is thus adopted as an empirical attackability estimator of $h$.
%In this sense, if the labels are strongly correlated in the training data set, the multi-label classifier learnt to capture the label correlation is prone to the non-targeted evasion attack. This is consistent with the discussions deduced from Theorem.\ref{bound:general} in Section.\ref{sec:adversarial_bound}.
%fuses both the intrinsic vulnerability and the attack strength together into one single score. The magnitude of the \textit{GASE} score 
%defines the alignment between the decision hyper-plane $\bf{w}_{i}$ and $\bf{w}_{j}$, measured by $cos\left\langle{-\bf{w}_{i}/h_{i}(\bf{x})},{-\bf{w}_{j}/h_{j}(\bf{x})}\right\rangle$. It

It is worth noting that \textbf{the proposed \textit{SAE} reflects the transferability of the attack, regardless of the setting of attack budget}. As shown by Eq.(\ref{eq:mattackr}), \textit{SAE} is evaluated only with the gradient information of the classifier, which is independent of the attack capability of the adversary. In contrast, \textit{GASE}  in \cite{zhuo2020} depends on the prior knowledge about the attack budget of the adversary. In practical applications, the attack budget is usually case-dependent, which limits the use of \textit{GASE} as a generic adversarial robustness evaluation tool. As an attack-strength-independent assessment, \textit{SAE} can help to evaluate the attackability level of a classifier, before it is compromised by any specific attack. It is therefore can be used as a predicative guide for choosing adversarially robust multi-label learning architectures. In the linear case where $h({x}) = {\bf{W}}{x}$, the cosine similarity $cos\left\langle{{\bf{a}}_j},{{\bf{a}}_k}\right\rangle$ produces a similar alignment metric as $ s(h_{j},h_{k}) = \max{\{cos\left\langle {{{\bf{w}}_j},{{\bf{w}}_k}} \right\rangle,}$ $ {cos\left\langle {{-{\bf{w}}_j},{{\bf{w}}_k}} \right\rangle\}}$. According to Eq.(\ref{eq:tbound}) and (\ref{eq:mattackc_v1}), the higher the cosine similarity score $cos\left\langle{{\bf{a}}_j},{{\bf{a}}_k}\right\rangle$ is, the higher $C_{{\bf{W}},z}$ in Eq.(\ref{eq:tbound}) and $A_{h({x}),\tilde{r}}$ in Eq.(\ref{eq:mattackc_v1}) becomes. %It demonstrates the use of $A_{h(\bf{x}),\tilde{r}}$ as an attackability estimator.  
%With the same magnitude $\varepsilon$ of the attack noise, the adversarial risk of $h$ increases as the better aligned decision boundaries, which is consistent with the observation derived from Eq.\ref{eq:mattackc}. 
%which then facilates the transfer of attack among labels and improve the attackability measured by ${A_{H({\bf{x}}),{\bf{r}}}}$.
% Given each label's attackability by $\bf{r}$ unchanged, bigger ${A_{H({\bf{x}}),{\bf{r}}}}$ indicates a more easily transferable attack. 
We thus measure the \textbf{attackability of $h$ at ${x}$} as the maximum $A_{h({x}),\tilde{r}}$ as: % which is conducted by looking for the optimal direction of $\bf{\tilde{r}}$, i.e
\begin{equation}\label{eq:attackt}
\small
{{\phi _{h,{{x}}}} = \mathop {\max }\limits_{{\tilde{r}}} {A_{h({{x}}),{{\tilde{r}}}}},}\,\,\,{s.t.\,\,\,{{\left\| {{\tilde{r}}} \right\|}_p} = 1}
\end{equation}
We inherit the constraint $\|\tilde{r}\|_{p}=1$ from Eq.(\ref{eq:linearize}). The resultant $\tilde{r}$ denotes the directions of the adversarial noise vector along which the attack can be maximally transferred. With this setting, we separate the derived transferability measurement with the attack strength.  
%\textcolor{blue}{Any discussion on the limited budget? ${\left\| {{\tilde{r}}} \right\|}_p = 1$}
%Specially, ${\phi _{h,{\bf{x}}}}$ evaluates the maximum number of labels $\bf{\tilde{r}}$ can flip with the constraint that $\bf{r}$ has unit $L_p$ norm.
With the primal-dual conversion, we can obtain the solution to Eq.(\ref{eq:attackt}) as:
\begin{equation}\label{eq:comb}
\small
\begin{array}{l}
{\phi _{h,{{x}}}} = \mathop {\max }\limits_{\{ {b_1},{b_2}, \cdot  \cdot  \cdot ,{b_m}\} } {\left\| {\sum\limits_{j = 1}^m {\frac{{ - {b_j}\nabla {h_j}({{x}})}}{{{h_j}({{x}})}}} } \right\|_q},\\
s.t.\,\,\,\,\frac{1}{p} + \frac{1}{q} = 1,\,\,\,{b_j} = \{ 0,1\} ,
\end{array}
\end{equation}
where $p$ denotes the $L_p$ norm of the perturbations. %The proof of Eq.(\ref{eq:comb}) is presented in supplementary document. 
%Our theory of SAE holds for all $l_p$ bounded adversaries when $p \ge 1$.
Without loss of generality, we only discuss $p=2$ of the $l_{p}$-norm in Eq.(\ref{eq:comb}). As the objective function of Eq.(\ref{eq:comb}) enjoys the submodularity property \cite{ElenbergAA2016}, we employ a simple yet effective greedy-based algorithm to solve Eq.(\ref{eq:comb}). Algorithm \ref{alg:estimator} describes the greedy-search based solution to compute the \textit{SAE} score. 
\begin{algorithm}	
	\DontPrintSemicolon
	\SetAlgoLined
	\caption{The Greedy Solution to Soft Attackability Estimation}
	\label{alg:estimator}
	\BlankLine
	\textbf{Input:}  $\left\{ {\frac{{ - \nabla {h_1}({{x}})}}{{{h_1}({{x}})}}, \cdots ,\frac{{ - \nabla {h_m}({{x}})}}{{{h_m}({{x}})}}} \right\}$. \\
	\BlankLine
	\textbf{Output:} The set of selected labels $S$. \\
	\BlankLine
	Initialize $S$ as an empty set. Set $LB=0$ and $CB=0$, where  $LB$ denotes the best result of last iteration and $CB$ denotes the best result of current iteration.\\
	\BlankLine
	\While{$|S|<m$}
	{
		$LB=CB$;\\
		$CB = \mathop {\max }\limits_{\left\{ {1, \cdots ,m} \right\} - S} \left( {\sum\limits_{i \in S} {\frac{{ - \nabla {h_i}({\bf{x}})}}{{{h_i}({\bf{x}})}} + \frac{{ - \nabla {h_j}({\bf{x}})}}{{{h_j}({\bf{x}})}}} } \right)$;\\
		if $CB<LB$, break;\\
		$S=S+j$
	}
\end{algorithm}

\subsection{SAE Regularized Multi-label Learning}
We propose to enhance the adversarial robustness of a multi-label classifier by enforcing the control over the \textit{SAE} score of the classifier explicitly during training.
%\textcolor{red}{what is $\alpha$ ? $\alpha > 0$? I don't understand why you introduce this setting.}
While we suppose ${x}$ is correctly classified during the theoretical analysis of attackability, it doesn't necessarily hold during training. For an originally miss-classified data instance ${x}$, it is possible that $A_{h(x),\tilde{r}}$ can be valued to 0. In this case, the attack perturbation $r$ can augment the confidence of the miss-classification. However, with $A_{h(x),\tilde{r}} = 0$, bare penalization can be enforced to suppress the bias. It may encourage further negative impact in the learnt classifier. 
%introduces bare penalization over the alignment between the decision hyper-planes and the parameters of the classifier $h$ (see Eq.\ref{eq:mattackc}). It may encourage miss-classification in the learnt classifier. 
To mitigate this issue, we slightly modify the definition of $A_{h(x),\tilde{r}}$ and use it as the transferability regularization term of multi-label learning, which gives:
\begin{equation}\label{eq:mlattackm}
\small
{\hat{A}_{h({{x}}),{{\tilde{r}}}}} = \sum\limits_{j = 1}^m {\max \{ \frac{{ - {{\tilde{r}}}^{T}{y_j}\nabla {h_j}({{x}})}}{{\max ({e^{{y_j}{h_j}({{x}})}},\alpha )}},0\} }  ,\,\,\,\alpha > 0
\end{equation}
%\textcolor{blue}{Using ${e^{{h_k}({\bf{x}}){y_k}}}$ helps assign more weights to the miss-classified data instances, thus gives more strength to regularize the decision boundaries to the positions classifying these instances correctly.}\textcolor{red}{I don't understand why it can drag the data instance towards the correct side.} 
% drag them to the correct side. 
where $\alpha$ is set to prevent over-weighing. For an originally correctly classified instance ($y_{j}h_{j}({x})>{0}$), $\hat{A}_{h(x),\tilde{r}}$ penalizes the attack transferability as $A_{h(x),\tilde{r}}$. For a miss-classified instance ($y_{j}h_{j}({x})\leq{0}$), minimizing $\hat{A}_{h(x),\tilde{r}}$ helps to reduce the confidence of the miss-classification output. Using the exponential function in $\hat{A}_{h(x),\tilde{r}}$, the miss-classified instance with stronger confidence (more biased decision output) is assigned with an exponentially stronger penalty. This setting strengthens the error-correction effect of $\hat{A}_{h(x),\tilde{r}}$.    

Similarly as in Eq.(\ref{eq:comb}), we can define ${\hat{\phi} _{h,{{x}}}} = \underset{\tilde{r}}{\max}{\hat{A}_{h(x),\tilde{r}}}$ in Eq.(\ref{eq:comb_aug}). The objective function of the SAE regularized multi-label learning (named hereafter as \textbf{ARM-SAE}) gives in Eq.(\ref{eq:loss}):
\begin{equation}\label{eq:comb_aug}
\small
\begin{array}{l}
{\hat{\phi} _{h,{{x}}}} = \mathop {\max }\limits_{\{ {b_1},{b_2}, \cdot  \cdot  \cdot ,{b_m}\} } {\left\| {\sum\limits_{j = 1}^m {\frac{{ - {b_j}{y_j}\nabla {h_j}({{x}})}}{{\max ({e^{{y_j}{h_j}({{x}})}},\alpha )}}}} \right\|_q},\\
s.t.\,\,\,\,\frac{1}{p} + \frac{1}{q} = 1,\,\,\,{b_j} = \{ 0,1\} ,
\end{array}
\end{equation}
 \begin{equation}\label{eq:loss}
l = \frac{1}{n}\sum_{i=1}^{n}\ell(h,z_{i}) + \frac{\lambda}{n}\sum_{i=1}^{n} {\hat{\phi}_{h,{{x}_{i}}}},
\end{equation} 
where %$z_{i} = \{(x_{i},y_{i})\}$ is the training data set and
$\lambda$ is the regularization weight. ${\hat{\phi} _{h,{{x}}}}$ can be calculated using the greedy search solution as ${\phi _{h,{{x}}}}$. If the classifier $h$ takes a linear form, we can find that \textit{ARM-SAE} reweighs the linear layer parameters of the classifier $\left\{ {{{\bf{w}}_{1,}} \cdots ,{{\bf{w}}_m}} \right\}$ with the weight $\frac{1}{\max ({e^{{y_j}{h_j}({{x}})}},\alpha )}$. Compared to \textit{ARM-Primal} (see Eq.(\ref{eq:comb}) to ${C_{{\bf{W}},z}}$ in Theorem \ref{bound:general}), \textit{ARM-SAE} enforces more transferability penalty over the instances with smaller classification margins. As unveiled in \cite{zhuo2020}, these instances are easier to be perturbed for the attack purpose. Instead of penalizing each instance with the same weight as in \textit{ARM-Primal}, \textit{ARM-SAE} can thus perform a more flexible instance-adapted regularization. %when training the multi-label classifier. 

\section{Experiments} \label{sec:exp}
%The goal of the experimental study is 3-fold: 
%\begin{itemize}
%	\item We verify the validity of the proposed empirical attackability estimator (\textit{SAE}).
%	\item We demonstrate the merit of the proposed \textit{ARM-SAE} method in improving adversarial robustness of the targeted multi-label classifier against evasion attack, compared to the state-of-the-art adversarial-robustness enhancing techniques for multi-label learning. 
%	\item We show the tunable trade-off between the utility of the classifier and its adversarial robustness by adjusting the regularization over the alignment between the decision hyperplanes of different labels.  
	
	%validating the trade-off between models' generalization performance on clean data and adversarial robustness, which is analyzed in Remark \ref{ob:trade}. 
% 	Given a multi-label learning problem, it's hard to change the inherent label dependency embedded in the data set. So, in this paper, we focus on the validation of the second kind of trade-off.
	%\item evaluating the effectiveness of proposed regularizer ARM-SAE. 
% 	Specially, we compare ARM to two well-established baselines and its' variants. 
%	Furthermore, we combine ARM with adversarial training to verify the further improvement of robustness.
%\end{itemize}
%1) validating the trade-off analysis in Obs. \ref{ob:trade}; and 2) evaluating the effectiveness of proposed regularizer MLAR.
\subsection{Experimental Setup}\label{exp:set}
\noindent{\textbf{Datasets.}}  
We empirically evaluate our theoretical study on three data sets collected from real-world multi-label cyber-security applications (\textit{Creepware}), object recognition (\textit{VOC2012}) \cite{pascal-voc-2012} and environment science (\textit{Planet}) \cite{planet2017}. The descriptions of the datasets are given can be found in the supplementary file due to the space limit. 
%including cyber security practices (\textit{Creepware}), object recognition (\textit{VOC2012}) \cite{pascal-voc-2012} and environment research (\textit{Planet}) \cite{planet2017}. 
%\textit{Creepware} data include different stalkware apps and each app has 16 labels indicating different types of surveillance on the victim's mobile device. 
%Besides, each app is profiled by the introductory texts of the app available in the app stores and signatures of its mobile service access. 
%\textit{VOC2012} is a well-known image data set and it is widely used in multi-label learning research.
%\textit{Planet} data collects daily satellite imagery of the entire land surface of the earth. Each image is equipped with labels denoting different atmospheric conditions and various classes of land cover/land use. 
The data sets are summarized in Table.\ref{tab:dataset}.

\noindent \textbf{Performance Benchmark.}\ \ Given a fixed attack strength of $\varepsilon$, we compute \emph{the number of flipped labels $C^{*}_h(z)$ on each testing instance according to Eq.(\ref{eq:attackdf})} and take the average of the derived $\{C^{*}_h(z)\}$ (noted as $C_{a}$) as an overall estimation of attackability on the testing data set. Due to the NP-hard intrinsic of the combinatorial optimization problem in Eq.(\ref{eq:attackdf}), we use \textit{GASE} \cite{zhuo2020} to estimate empirically $C^{*}_h(z)$ and $C_{a}$. 
% We gradually increase the attack budget $\varepsilon$ and plot the $\varepsilon  - {C_a}$ curve to demonstrate the variation of the attackability level with increasingly stronger attack. 
Besides, we measure the multi-label classification performance on the clean and adversarially modified testing instances with \textit{Micro-F1} and \textit{Macro-F1} scores.  
%Then the area under curve (AUC) of the obtained $\varepsilon  - {C_a}$ curve is calculated as the metric to comprehensively evaluate a model's attackability with varying attack budget. Usually, we plot a \emph{classification performance on clean data $-$ estimated attackability} curve to show the trade-off between the utility of the classifeir and its adversarial robustness. 
%For short, we denote the \emph{classification performance on clean data $-$ estimated attackability} curve as ${C_{clean}} - AU{C_\varepsilon }$ curve.

%\vspace{5 pt}
\noindent{\textbf{Targeted Classifiers.}} 
We instantiate the study empirically with linear Support Vector Machine (SVM) and Deep Neural Nets (DNN) based multi-label classifiers.  
Linear SVM is applied on \textit{Creepware}.     DNN model Inception-V3 \cite{inceptionv3} is used on \textit{VOC2012} and \textit{Planet}. On each data set, we randomly choose $50\%$, $30\%$ and $20\%$ data instances for training, validation and testing to build the targeted multi-label classifier.  In Table.\ref{tab:dataset}, we show \textit{Micro-F1} and \textit{Macro-F1} scores measured on the clean testing data to evaluate the classification performance of multi-label classifiers \cite{mmf}.
{Note that accurate adversary-free multi-label classification is beyond our scope and these classifiers are used to verify the theoretical analysis and the proposed \textit{ARM-SAE} method.}

\begin{table}[t!]
	\small                    % change according to needs
	\setlength\tabcolsep{2.0pt} % change according to needs
	\caption {Summary of the used real-world  data sets. %Each classifier is trained and tested on ${\rm{80\% }}$ and ${\rm{20\% }}$ instances of each database respectively. 
		$N$ is the number of instances. $m$ is the total number of labels. $ {l}_{avg}$ is the average number of labels per instance.
		The F1-scores of the targeted classifiers on different data sets are also reported. }	 
	\label{tab:dataset}
	\newcommand{\tabincell}[2]{\begin{tabular}{@{}#1@{}}#2\end{tabular}}
	\centering
	\begin{tabular}{c|c|c|c|c|c|c} 
		\hline
		\hline
		Data set	& $N$ & m & ${l}_{avg}$ & Micro F1 & Macro F1 & Classifier$_{target}$\\
		\hline
		{Creepware}	&966	&16 & 2.07   &0.76  & 0.66   & SVM		\\
		\hline
		{VOC2012}	&17,125 	&20 & 1.39  & 0.83 &0.74  & Inception-V3  \\
		\hline
		{Planet}	&40,479 	&17 & 2.87   & 0.82  &0.36   & Inception-V3   \\
		\hline
		\hline
	\end{tabular}
\end{table}

%\vspace{5 pt}
%\noindent {\textbf{Hard Empirical Attackability Estimator (GASE \cite{zhuo2020}).}}\ \ We follow the definition in Eq.(\ref{eq:attackdf}) to evaluate the hard empirical attackability of a multi-label classifier. Unfortunately, Eq.(\ref{eq:attackdf}) raises up a  NP-hard mixed-integer non-linear constraint problem (MINLP) and solving it is not the focus of this paper. Paper \cite{zhuo2020} proposed an effective algorithm named GASE to estimate $C^*$ in Eq.(\ref{eq:attackdf}) and we also adopt it as the hard attackability estimator in this paper. In short, GASE greedily expand the attack space by adding the label with the largest marginal gain to the attack sequence. 
% Note that \textit{projected gradient decent} (PGD) \cite{Madry2018iclr} is employed to conduct the targeted attack in GASE. 

%\vspace{5 pt}

%\vspace{5 pt}
\noindent {\textbf{Input Normalization and Reproduction.}}\ \  
We normalize the adversarially perturbed data during the attack process. Due to the space limit, we provide the parameter settings and the reproduction details in the supplementary file. 
%When imposing attacks, we project the perturbed data in  \emph{VOC2012} and \emph{Planet} to $[ - 1,1]$, while we don't limit the value range of data in \emph{Creepware}. The $\alpha$ in Eq.(\ref{eq:loss}) is empirically set to 0.01 in all experiments. The regularization parameters $\lambda$ in Eq.\ref{eq:loss} and other baselines are chosen empirically from the range $\left\{ {{{10}^{ - 8}},{{10}^{ - 7}}, \cdots ,{{10}^7},{{10}^8}} \right\}$
% The attack budget used to calculate the $AUC$ score  is setted on each data as: \emph{Creepware (0.5), VOC2012 (10)} and \emph{Planet (4)}. For an intuition, a budget of $10$ on data set \emph{VOC2012} allows changeing $25$ pixels by 255. 

%Our codes were written in Python and all the models were built by Keras package \cite{chollet2015keras}. The needed targeted evasion attack and adversarial training are implemented by adversarial-robustness-toolbox \cite{art2018toolbox}. Our experiments were conducted on GPU rtx2080ti. 
% The regularization parameter $\lambda$ in Eq.\ref{eq:loss} is choosen empirically from the range $\{1e-2,1e-1,1,1e1,1e2,1e3\}$ 
% \textcolor{red}{this is an example, you can put the true lambda values that you have traversed}.

\subsection{Effectivity of Soft Attackability Estimator (SAE)}

\begin{table*}[!htb]
	\small                    % change according to needs
	\scriptsize 
	\setlength\tabcolsep{3pt} % change according to needs
	\caption{Attackability estimation by \textit{SAE}. ${\lambda _{nuclear}}$ denotes the strength of nuclear-norm based regularization. $CC$ and $P$ denote the Spearman coefficient and the p-value between \textit{GASE} and \textit{SAE} scores on the testing instances. } 
	\label{result:SAE}
	\begin{center}
		\renewcommand\arraystretch{1}
		\resizebox{\textwidth}{!}{ 
			\begin{tabular}{c|c|c c c c c|c}
				\hline
				\hline
				Data set& & \multicolumn{5}{c|}{$ \longrightarrow robustness\,\,increase$ }  & $CC,P(Spearman)$\\
				\cline{1-8} 
				\multirow{3}{*}{\textit{Creepware}} &${\lambda _{nuclear}}$     & 0 & 0.00001 & 0.0001 & 0.001 & 0.01 & \multirow{3}{*}{$\begin{array}{l}
					CC = 1\\
					P = 0
					\end{array}$} \\
				&GASE (${C_a},\varepsilon  = 0.5$)    & 13.5& 11.4& 10.8& 6.9& 4.3  & \\
				&SAE     & 31.5& 19.16& 18.06& 14.55& 11.22 &  \\
				\hline			

%				&\multicolumn{5}{c|}{\textbf{$VOC2012$} }  & $CC,P(Spearman)$ \\
%				\cline{1-7} 
				\multirow{3}{*}{\textit{VOC2012}} &${\lambda _{nuclear}}$     & 0 & 0.0001 & 0.001 & 0.01 & 0.1 
				& \multirow{3}{*}{$\begin{array}{l}
					CC = 1\\
					P = 0
					\end{array}$} \\
				&GASE (${C_a},\varepsilon  = 10$)    & 10.8& 10.1& 9.3& 8.5& 4.9 &  \\
				&SAE     & 157.6& 127.3& 77.6& 69.1& 61.0 &  \\
				\hline

%				&\multicolumn{5}{c|}{\textbf{$Planet$} }  & $CC,P(Spearman)$ \\
%				\cline{1-7} 
				\multirow{3}{*}{\textit{Planet}} &${\lambda _{nuclear}}$     & 0 & 0.0001 & 0.001 & 0.01 & 0.1 
				& \multirow{3}{*}{$\begin{array}{l}
					CC = 1\\
					P = 0
					\end{array}$} \\
				&GASE (${C_a},\varepsilon  = 2$)    & 13.1& 12.2& 11.6& 10.5& 7.1 &  \\
				&SAE     & 267.1& 221.5& 186.3& 158.2& 102.0 &\\
				\hline
				 &&\multicolumn{5}{c|}{$ \longrightarrow attackability\,\,decrease$  }   &\\
				\hline
				\hline
			\end{tabular}
		}
	\end{center}
\end{table*}

%computed from a given multi-label classifier

In Table.\ref{result:SAE}, we demonstrate the validity of the proposed \textit{SAE} by checking the consistency between the \textit{SAE} and the \textit{GASE}-based attackability measurement \cite{zhuo2020}. We adopt the \textbf{nuclear-norm} regularized training \cite{zhuo2020} to obtain an adversarially robust multi-label classifier. On the same training set, we increase the nuclear-norm regularization strength gradually to derive more robust architectures against the evasion attack. For each regularization strength, we can compute the \textit{SAE} score of the classifier on the unperturbed testing instances. Similarly, by freezing the attack budget $\varepsilon$ on each data set, we can generate the \textit{GASE} score ($C_a$) corresponding to each regularization strength. Note that \textit{only the ranking orders of the \textit{SAE} and \textit{GASE} score matters in the attackabiltiy measurement by definition}. We use the ranking relation of the scores to select adversarially robust models. Therefore, we adopt the Spearman rank correlation coefficient to measure the consistency between \textit{SAE} and \textit{GASE}. 

We use the \textit{GASE} score as a baseline of attackability assessment. The \textit{SAE} and \textit{GASE} score are strongly and positively correlated over all the datasets according to the correlation metric. 
% Though $p>0.05$ for the Pearson coefficient on \textit{VOC2012}, \textit{SAE} and \textit{GASE} score presents a clear \textit{monotonic relation} on the dataset. The value of one score increases/decreases, then so does the other's value, but not exactly at a constant rate. The monotonic relation indicates the consistency between \textit{SAE} and \textit {GASE}. Note that the Pearson coefficient only assesses the linear relation. It doesn't capture the monotonic relation shown on \textit{VOC2012}. In contrast, 
% The Spearman coefficient  does reflect a significantly strong monotonic correlation. 
Furthermore, with a stronger robustness regularization, the \textit{SAE} score decreases accordingly. It confirms that the intensity of the proposed \textit{SAE} score capture the attackability level of the targeted classifeir. This observation further validates empirically the motivation of using \textit{SAE} in adjusting the adversarial robustness of the classifier. 

The experimental study also shows the attack-strength-independent merit of the \textit{SAE} over \textit{GASE}. \textit{SAE} is computed without knowing the setting of the attack budget. It thus reflects the intrinsic property of the classifier determining its adversarial vulnerability. In practice, this attack-strength-independent assessment can help to evaluate the attackability level of the deployed classifier, before it is compromised by any specific attack.

%With increasingly stronger attack (larger $\varepsilon$), the \textit{GASE} score rises accordingly, which indicates the increase of the attackability level. The value of $\textit{SAE}$ presents a highly similar variation tendency as the \textit{GASE} score. It confirms that the intensity of the proposed \textit{SAE} score capture the attackability level of the targeted classifeir. This observation further validates empirically the motivation of using \textit{SAE} in adjusting the adversarial robustness of the classifier. 

%the hard empirical attackability of a multi-label classifier. Unfortunately, Eq.(\ref{eq:attackdf}) raises up a  NP-hard mixed-integer non-linear constraint problem (MINLP) and solving it is not the focus of this paper. Paper \cite{zhuo2020} proposed an effective algorithm named GASE to estimate $C^*$ in Eq.(\ref{eq:attackdf}) and we also adopt it as the hard attackability estimator in this paper.

\subsection{Effectiveness Evaluation of ARM-SAE}
We compare the proposed \textit{ARM-SAE} method to the state-of-the-art techniques in improving adversarial robustness of multi-label learning: the $L_{2}$-norm and the nuclear-norm regularized multi-label training \cite{zhuo2020}. Besides, we conduct an ablation study to verify the effectiveness of \textit{ARM-SAE}.
\begin{itemize}
	\item \textbf{$L_2$ Norm and Nuclear Norm Regularized Training.}  
%	$L_2$ regularizer is widely used to control the $L_2$ norm of a model's weights. 
    Enforcing the $L_2$ and nuclear norm constraint 
    % to encourage a low-rank structure of the classifier's parameters 
    helps to reduce the model complexity and thus enhance the model's adversarial robustness \cite{TuNIPS19,zhuo2020}. 
%	Besides, for a linear SVM classifier, the reduced $L_2$ norm means a wider margin, which is supposed to imply a better robustness \cite{margin}. 
	%\item \textbf{Nuclear-norm regularized training.} In \cite{zhuo2020}, 
%	a low-rank structure of weight matrix is proved to be favorable to the adversarial robustness of a multi-label classifier. Then 
	%A nuclear-norm based constraint is enforced to encourage low rank structures of the classifier's parameters. 
	\item \textbf{ARM-Single.} This variant of \textit{ARM-SAE} is built by enforcing the transferability regularization with respect to individual labels separately:
	\begin{equation}\label{eq:single}
	\small
	{\phi _{H\_{\rm{single}}}} = \sum_{i=1}^{n} {\sum\nolimits_{k = 1}^m {{{\left\| {\frac{{\nabla {h_k}({{x_{i}}})}}{{\max ({e^{{y^{i}_k}{h_k}({{x}})}},\alpha )}}} \right\|}_2}} } \,.
	\end{equation}
	We compare \textit{ARM-SAE} with \textit{ARM-Single} to show the merit of jointly measuring and regularizing the impact of the input attack noise over all the labels. 
	\textit{ARM-SAE} tunes the transferability of the attack jointly, while \textit{ARM-Single} enforces the penalization with respect to each label individually.  
% 	The design of \textit{ARM-SAE} penalizes the classifier's architecture of each label in a self-adaptive way, by considering the transferability of the attack.	In contrast, the label-wise penalization can overweight the penalization over some labels, while underweight the others. 
	
	%ARM can inhibit the transfer of attack. 
	\item \textbf{ARM-Primal.} We compare this variant to \textit{ARM-SAE} to demonstrate the merit of \textit{ARM-SAE} by 1) introducing the flexiblity of penalizing the whole model architecture, instead of only the linear layer; 2) taking the impact of classification margin on adversarial risk \cite{margin} into the consideration. 
	
	\begin{table*}[!htb]
	%	\small                    % change according to needs
%	\scriptsize 
	\setlength\tabcolsep{3pt} % change according to needs
	\caption{Effectiveness evaluation of ARM-SAE. For convenience, \textit{non}, $L_2$, \textit{nl}, \textit{sg}, \textit{pm} and \textit{SAE} are used to denote the absence of regularization, \textit{$L_2$ norm, nuclear-norm, ARM-single, ARM-Primal} and \textit{ARM-SAE} based methods respectively. The best results are in bold.}
	\label{result:ARM}
	\begin{center}
		\renewcommand\arraystretch{1}
		\resizebox{\textwidth}{!}{ 
			\begin{tabular}{c|c c c c c c |c c c c c c}
				\hline
				\hline
				&\multicolumn{12}{c}{\textbf{$Creepware:$ }Micro F1 = 0.76, Macro F1 = 0.66 (on clean data) }  \\
				\cline{1-13} 
				Budget	&\multicolumn{6}{c}{\textbf{$\varepsilon  = 0.05$} } &\multicolumn{6}{|c}{\textbf{$\varepsilon  = 0.2$} }\\	
				\cline{1-13} 		
				Regularizors     & \textit{non} & $L_2$ & \textit{nl} & \textit{sg} & \textit{pm} & \textit{SAE} & \textit{non} & $l_2$ & \textit{nl} & \textit{sg} & \textit{pm} & \textit{SAE} \\
				Micro F1    & 0.34& 0.40 &0.45& 0.44&  0.43& \textbf{0.53} & 0.10& 0.13& 0.15& 0.15& 0.16& \textbf{0.22} \\
				Macro F1     & 0.33& 0.39 &\textbf{0.43}& 0.39&  \textbf{0.43}& 0.42 & 0.12& 0.15& 0.20& 0.17& 0.20& \textbf{0.25}   \\
				\hline			

				\hline
				&\multicolumn{12}{c}{\textbf{$VOC2012:$} Micro F1 = 0.83, Macro F1 = 0.74 (on clean data)}  \\
				\cline{1-13} 
				Budget	&\multicolumn{6}{c}{\textbf{$\varepsilon  = 0.1$} } &\multicolumn{6}{|c}{\textbf{$\varepsilon  = 1$} }\\	
				\cline{1-13}		
				Regularizors     & \textit{non} & $L_2$ & \textit{nl} & \textit{sg} & \textit{pm} & \textit{SAE} & \textit{non} & $l_2$ & \textit{nl} & \textit{sg} & \textit{pm} & \textit{SAE} \\
				Micro F1    & 0.49& 0.53& 0.56& 0.54& 0.57& \textbf{0.61} & 0.20& 0.22& 0.27& 0.26& 0.26& \textbf{0.30} \\
				Macro F1    & 0.29& 0.31& 0.33& 0.31& 0.36& \textbf{0.38} & 0.12& 0.16& 0.22& 0.17& 0.20& \textbf{0.23} \\
				\hline
				
				\hline
				&\multicolumn{12}{c}{\textbf{$Planet:$} Micro F1 = 0.82, Macro F1 = 0.36 (on clean data)}  \\
				\cline{1-13} 
				Budget	&\multicolumn{6}{c}{\textbf{$\varepsilon  = 0.1$} } &\multicolumn{6}{|c}{\textbf{$\varepsilon  = 1$} }\\	
				\cline{1-13}		
				Regularizors     & \textit{non} & $L_2$ & \textit{nl} & \textit{sg} & \textit{pm} & \textit{SAE} & \textit{non} & $l_2$ & \textit{nl} & \textit{sg} & \textit{pm} & \textit{SAE} \\
				Micro F1    & 0.41& 0.49& 0.45& 0.48& 0.49& \textbf{0.53} & 0.06& 0.09& 0.08& 0.10.& 0.09& \textbf{0.13} \\
				Macro F1     & 0.13& 0.22& 0.17& 0.20& 0.18& \textbf{0.24} & 0.03& 0.04& 0.04& 0.06& 0.06& \textbf{0.08} \\
				\hline
				\hline
			\end{tabular}
		}
	\end{center}
\end{table*}

\begin{table*}[!htb]
	%	\small                    % change according to needs
	\scriptsize 
	\setlength\tabcolsep{3pt} % change according to needs
	\caption{Trade-off Between Generalization Performance on Clean Data and Adversarial Robustness on \textit{Creepware}. The attack budget $\varepsilon=0.05$.}
	\label{result:trade}
	\begin{center}
		\renewcommand\arraystretch{1}
		\resizebox{0.7\textwidth}{!}{ 
			\begin{tabular}{c|c c c c c}
				\hline
				\hline
				$\lambda$     & 0 & ${10^{ - 7}}$ & ${10^{ - 6}}$ & ${10^{ - 5}}$ & ${10^{ - 4}}$   \\
				\cline{1-6} 
				${\phi _{align}}$     & 0.23 & 0.22 & 0.20 & 0.15 & 0.12   \\
				Micro F1(clean)    & 0.76& 0.76& 0.75& 0.72& 0.70   \\
				Macro F1(clean)     & 0.66& 0.63& 0.56& 0.50& 0.46    \\
				MIcro F1(pert)    & 0.34& 0.35& 0.39& 0.44& 0.53  \\
				Macro F1(pert)     & 0.33& 0.33& 0.35& 0.40& 0.42   \\
				\hline			
				\hline
			\end{tabular}
		}
	\end{center}
\end{table*}

\end{itemize}
Two different attack budgets $\varepsilon$ on each data set are introduced denoting varied attack strength. With each fixed $\varepsilon$, we compute the \textit{Micro-F1} and \textit{Macro-F1} scores of the targeted classifiers after retraining with the techniques above. Table.\ref{result:ARM} lists the classification accuracy over the adversarial testing instances using different robust training methods. In Table.\ref{result:ARM}, we also show the multi-label classification accuracy (measured by two F1 scores) on the clean testing instances as a baseline. 
Consistently observed on the three datasets, even a small attack budget can deteriorate the classification accuracy drastically, which shows the vulnerability of multi-label classifiers. Generally, all the regularization method can improve the classification accuracy on the adversarial input. Among all the methods, \textit{ARM-SAE} achieves the highest accuracy on the adversarial samples. It confirms the merit of \textit{SAE} in controlling explicitly the transferability and then suppressing the attackabilty effectively. In addition, by regularizing jointly the attack transfer and exploiting classification margin for the attackability measurement, \textit{ARM-SAE} achieves superior robustness over the two variants. %\textit{ARM-Single} and \textit{ARM-Primal}.   

\subsection{Validation of Trade-off Between Generalization Performance on Clean Data and Adversarial Robustness}

We validate the trade-off described in Remark.\ref{ob:trade}. Without loss of generality, we conduct the case study on \textit{Creepware}. Tuning the alignment between decision boundaries of different labels is achieved by conducting the \textit{ARM-SAE} training as described in Eq.\ref{eq:loss}. %Eq.(\ref{eq:align}). 
%\begin{equation}\label{eq:align} 
%l = \frac{1}{n}\sum\limits_{i = 1}^n \ell  (h,{z_i}) + \frac{\lambda }{n}\sum\limits_{i = 1}^n {\sum\limits_{j,k \in \{ 1, \cdots ,m\} } {\cos \left\langle {y_j^i\nabla {h_j}(x),\left. {y_k^i\nabla {h_k}(x)} \right\rangle } \right.} }. 
%\end{equation}
%Note that compared to ARM-SAE, Eq.(\ref{eq:align}) filters out the regularization on the $L2$ norm of weights, which ensures that the improvement of adversarial robustness comes form the adjustment of alignment between decision boundaries, but not from the reducement of weight complexity. 
We freeze $\varepsilon$ as 0.05 and vary the regularization weight $\lambda$ in Eq.(\ref{eq:loss}) from ${10^{ - 7}}$ to ${10^{ - 4}}$ to show increasingly stronger regularization effects enforced on the alignment between decision boundaries of different labels. For each regularization strength, we train a multi-label classifier $h$ and evaluate quantitatively the averaged alignment level ${\phi _{align}} = \frac{1}{{{m^2}}}\sum\limits_{j,k \in \{ 1, \cdots ,m\} } {\left| {\cos \left\langle {{{\bf{w}}_j},\left. {{{\bf{w}}_k}} \right\rangle } \right.} \right|} $ between the decision hyperplanes of different labels. Table.\ref{result:trade} shows the variation of $\phi_{align}$ and the Micro- / Macro-F1 accuracy of the trained multi-label classifier $h$ over the clean and adversarially perturbed data instances (Micros / Macro F1 (clean / pert)). With increasingly stronger robustness regularization, the averaged alignment level $\phi_{align}$ between the label-wise decision hyper-planes decreases accordingly. Simultaneously, we witness the rise of the classification accuracy of $h$ on the adversarially perturbed testing instances. It indicates the classifier $h$ is more robust to the attack perturbation. However, the Macro- and Micro-F1 scores of $h$ on the clean testing data drop with stronger alignment regularization. This observation is consistent with the discussion in Remark.\ref{ob:trade}.

\section{Conclusion}\label{sec:conclusion}
In this paper, we establish an information-theoretical adversarial risk bound of multi-label classification models. Our study identifies that the transferability of evasion attack across different labels determines the adversarial vulnerability of the classifier. Though capturing the label correlation improves the accuracy of adversary-free multi-label classification, our work unveils that it can also encourage transferable attack, which increases the adversarial risk. We show that the trade-off between the utility of the classifier and its adversarial robustness can be achieved by explicitly regularizing the transferability level of evasion attack in the learning process of multi-label classification models. Our empirical study demonstrates the applicability of the proposed transferability-regularized robust multi-label learning paradigm for both linear and non-linear classifies.   
%Empirical study on the proposed transferability-regularized multi-label learning paradigm confirms the merits of suppressing attack transfer to gain adversarial robustness in multi-label learning scenarios. 

%The resultant transferability-tuning based robust multi-label learning 
% We observe that strong label dependency facilitates the transfer of attack by approaching the decision hyperplanes closer. 
%Based on the theoretical analysis, we develop a trasnferability regularizor to improve the adversarial robustness of multi-label classifiers, especially by inhibiting the transfer of attack. 
% ARM is based on a soft empirical attackability measurement method (SAE), which captures the attack transferability effectively. 
%Our theoretical analysis about the trade-off between the generalization on clean data and adversarial robustness and the effectiveness of ARM is verified by the organized experiments. 
%the key factors characterizing the vulnerability of the classifier: 1) the conditional mutual information between the training data and trained model, 2) the correlation among labels and the learned classifiers and 3) attack strength mainly determine the attackability of a multi-label model.

\bibliographystyle{splncs04}
\bibliography{reference}

\begin{thebibliography}{10}
\providecommand{\url}[1]{\texttt{#1}}
\providecommand{\urlprefix}{URL }
\providecommand{\doi}[1]{https://doi.org/#1}

\bibitem{chollet2015keras}
Chollet, Francois: Keras (2015), \url{https://github.com/fchollet/keras}

\bibitem{ElenbergAA2016}
Elenberg, E.R., Khanna, R., Dimakis, A.G., Negahban, S.: Restricted strong
  convexity implies weak submodularity. Annuals of Statistics  (2016)

\bibitem{margin2}
Elsayed, G.F., Krishnan, D., Mobahi, H., Regan, K.: Large margin deep networks
  for classification. In: NeuIPS (2018)

\bibitem{pascal-voc-2012}
Everingham, M., Gool, L.V., Williams, C., Winn, J., Zisserman, A.: The pascal
  visual object classes challenge 2012 (voc2012) results.
  \url{http://www.pascal-network.org/challenges/VOC/voc2012/workshop/index.html"}
  (2012)

\bibitem{AFawzi2016nips}
Fawzi, A., Moosavi-Dezfooli, S., Frossard, P.: Robustness of classifiers: From
  adversarial to random noise. In: NIPS. p. 1632–1640 (2016)

\bibitem{Fawzi2020MachineLearning}
Fawzi, A., Fawzi, O., Frossard, P.: Analysis of classifiers' robustness to
  adversarial perturbations. Machine Learning  \textbf{107},  481--508 (2018)

\bibitem{Ristenpart2018chi}
Freed, D., Palmer, J., Minchala, D., Levy, K., Ristenpart, T., Dell, N.: “a
  stalker's paradise”: How intimate partner abusers exploit technology. In:
  the 2018 CHI Conference. p. 1–13 (2018)

\bibitem{Gilmer2018arxiv}
Gilmer, J., Metz, L., Faghri, F., Schoenholz, S., Raghu, M., Wattenberg, M.,
  Goodfellow, I.: Adversarial spheres. CoRR  (2018),
  \url{http://arxiv.org/abs/1801.02774}

\bibitem{Gupta2013www}
Gupta, A., Lamba, H., Kumaraguru, P., Joshi, A.: Faking sandy: Characterizing
  and identifying fake images on twitter during hurricane sandy. In: WWW. p.
  729–736 (2013)

\bibitem{Matthias2017nips}
Hein, M., Andriushchenko, M.: Formal guarantees on the robustness of a
  classifier against adversarial manipulation. In: NeuIPS. pp. 2266--2276
  (2017)

\bibitem{hoe63}
Hoeffding, W.: Probability inequalities for sums of bounded random variables.
  Journal of the American Statistical Association  \textbf{58}(301),  13--30
  (1963)

\bibitem{planet2017}
Kaggle: Planet: Understanding the amazon from space.
  \url{https://www.kaggle.com/c/planet-understanding-the-amazon-from-space/overview}
  (2017)

\bibitem{JKhim2018ft}
Khim, J., Loh, P.L.: Adversarial risk bounds for binary classification via
  function transformation. arXiv  (2018)

\bibitem{art2018toolbox}
Nicolae, M., Sinn, M., Minh, T.N., Rawat, A., Wistuba, M., Zantedeschi, V.,
  Molloy, I.M., Edwards, B.: Adversarial robustness toolbox v0.2.2. CoRR
  (2018), \url{http://arxiv.org/abs/1807.01069}

\bibitem{kevin2020sp}
Roundy, K.A., Mendelberg, P.B., Dell, N., McCoy, D., Nissani, D., Ristenpart,
  T., Tamersoy, A.: The many kinds of creepware used for interpersonal attacks.
  In: IEEE Symposium on Security and Privacy (SP). pp. 626--643 (may 2020)

\bibitem{SongQi2018ICDM}
Song, Q., Jin, H., Huang, X., Hu, X.: Multi-label adversarial perturbations.
  In: ICDM. pp. 1242--1247 (2018)

\bibitem{cmi2020}
Steinke, T., Zakynthinou, L.: Reasoning about generalization via conditional
  mutual information. In: COLT (2020)

\bibitem{inceptionv3}
Szegedy, C., Vanhoucke, V., Ioffe, S., Shlens, J., Wojna, Z.: Rethinking the
  inception architecture for computer vision. In: arXiv (2015)

\bibitem{TuNIPS19}
Tu, Z., Zhang, J., Tao, D.: Theoretical analysis of adversarial learning: {A}
  minimax approach. In: NeuIPS. pp. 12259--12269 (2019)

\bibitem{Wang2018AnalyzingTR}
Wang, Y., Jha, S., Chaudhuri, K.: Analyzing the robustness of nearest neighbors
  to adversarial examples. In: ICML. pp. 5133--5142 (2018)

\bibitem{Han2020kdd}
Wang, Y., Han, Y., Bao, H., Shen, Y., Ma, F., Li, J., Zhang, X.: Attackability
  characterization of adversarial evasion attack on discrete data. In: KDD. pp.
  1415--1425 (2020)

\bibitem{margin}
Yang, Y., Khanna, R., Yu, Y., Gholami, A., Keutzer, K., Gonzalez, J.E.,
  Ramchandran, K., Mahoney, M.W.: Boundary thickness and robustness in learning
  models. In: NeuIPS (2020)

\bibitem{YangNeuIPS2020}
Yang, Y., Rashtchian, C., Zhang, H.: A closer look at accuracy vs. robustness.
  In: NeuIPS (2020)

\bibitem{zhuo2020}
Yang, Z., Han, Y., Zhang, X.: Characterizing the evasion attackability of
  multi-label classifiers. In: AAAI (2021)

\bibitem{Yin2018icml}
Yin, D., Ramchandran, K., Bartlett, P.L.: Rademacher complexity for
  adversarially robust generalization. In: ICML. pp. 7085--7094 (2019)

\bibitem{zhang2019icml}
Zhang, H., Yu, Y., Jiao, J., Xing, E.P., Ghaoui, L.E., Jordan, M.I.:
  Theoretically principled trade-off between robustness and accuracy. In: ICML
  (2019)

\bibitem{mmf}
Zhang, M., Zhou, Z.: A review on multi-label learning algorithms. TKDE
  \textbf{26}(8),  1819--1837 (2013)

\bibitem{zhao2019aaai}
Zhao, C., Fletcher, P., Yu, M., Peng, Y., Zhang, G., Shen, C.: The adversarial
  attack and detection under the fisher information metric. In: AAAI. pp.
  5869--5876 (2019)

\end{thebibliography}

\clearpage 
\section{Supplementary}
\subsection{Database Summary}
We empirically evaluate our theoretical study on three data sets collected from real-world multi-label applications. They include cyber security practices (\textit{Creepware}), object recognition (\textit{VOC2012}) \cite{pascal-voc-2012} and environment research (\textit{Planet}) \cite{planet2017}. 
\textit{Creepware} data include different stalkware app instances and each instance has 16 labels indicating different types of surveillance on the victim's mobile device. 
Besides, each app is profiled by the introductory texts of the app available in the app stores and signatures of its mobile service access. 
\textit{VOC2012} is a well-known image data set and it is widely used in multi-label learning research.
\textit{Planet} data collects daily satellite imagery of the entire land surface of the earth. Each image is equipped with labels denoting different atmospheric conditions and various classes of land cover/land use. %The data sets are summarized in Table.\ref{tab:dataset}.

\subsection{Input Normalization and Parameter settings}
When imposing attacks, we project the perturbed data in  \emph{VOC2012} and \emph{Planet} to $[ - 1,1]$, while we don't limit the value range of data in \emph{Creepware}. The $\alpha$ in Eq.(\ref{eq:loss}) is empirically set to 0.01 in all experiments. The regularization parameters $\lambda$ in Eq.\ref{eq:loss} and other baselines are chosen empirically from the range $\left\{ {{{10}^{ - 8}},{{10}^{ - 7}}, \cdots ,{{10}^7},{{10}^8}} \right\}$
% The attack budget used to calculate the $AUC$ score  is setted on each data as: \emph{Creepware (0.5), VOC2012 (10)} and \emph{Planet (4)}. For an intuition, a budget of $10$ on data set \emph{VOC2012} allows changeing $25$ pixels by 255. 

Our codes were written in Python and all the models were built by Keras package \cite{chollet2015keras}. The needed targeted evasion attack and adversarial training are implemented by adversarial-robustness-toolbox \cite{art2018toolbox}. Our experiments were conducted on GPU rtx2080ti. Our codes for SAE training are available at \url{https://github.com/chelungungun/Transferability_MLATTACK} 

\subsection{Proofs}
We supple the proof of the theorem in our paper, especially the Eq.(\ref{eq:tbound}) and Eq.(\ref{eq:cmibound}), and the proof from Eq.(\ref{eq:attackt}) to Eq.(\ref{eq:comb}) 

\begin{lemma}\label{le:ex}
	(Thomas 2020 \cite{cmi2020}, Corollary 5) Let $E$, $E'$ and $Z$ be independent random variables where $E$ and $E'$ have identical distributions. Let $A$ be a random function whose randomness is independent from $E$, $E'$ and $Z$. Let $g$ be a fixed function. Then
	\begin{equation}
	\begin{array}{l}
	\mathop \mathbb{E}\limits_{A,E,Z} \left[ {g(A(E,Z),E,Z)} \right]\\
 	\le \mathop {\inf }\limits_{t > 0} \frac{{I(A(E,Z);E|Z) + \mathop \mathbb{E}\limits_Z \left[ {\log \mathop \mathbb{E}\limits_{A,E,E'Z} \left[ {{e^{t \cdot g(A(E,Z),E',Z)}}} \right]} \right]}}{t}
	\end{array}
	\end{equation}
\end{lemma}

\begin{lemma}\label{le:in}
	(Hoeffding 1963 \cite{hoe63}.) Let ${\bf{X}} \in \left[ {a,b} \right]$ be a random variable with mean $\mu $. Then for all $t \in \mathbb{R}$,
	\begin{equation}
	\mathbb{E}({e^{t{\bf{X}}}}) \le {e^{t\mu  + {t^2}{{(b - a)}^2}/8}}
	\end{equation}
\end{lemma}

\textbf{Proof from Eq.\ref{eq:attackt} to Eq.\ref{eq:comb}:}
We can rewrite Eq.\ref{eq:mattackr} as
\begin{equation}\label{eq:mattackr2}
{A_{h({{x}}),{\tilde{r}}}} = {\tilde{r'}}\sum\limits_{k = 1}^m {\frac{{ - \nabla {h_k}({{x}}) * \max \{ {\rm{sgn}}( - {\tilde{r'}}{y_k}\nabla {h_k}({{x}})),0\} }}{{{h_k}({{x}})}}}. 
\end{equation}
If there is no sgn function and max function in Eq.\ref{eq:mattackr2}, Eq.\ref{eq:attackt} is actually the definition of dual norm. To eliminate the sgn and max function,  we can break the domain of $\tilde{r}$ into a group of subsets according the output of those sgn functions. Denote the domain of $\tilde{r}$ as $I$ and $I_S$ is a subset of $I$ which is defined by Eq.(\ref{eq:subdomain}). $S$ is an element from the power set of $\left\{ {1, \cdot  \cdot  \cdot ,m} \right\}$. 
\begin{equation}\label{eq:subdomain}
{I_S} = \left\{ {{\tilde{r}} \left| {\begin{array}{*{20}{c}}
		{{\tilde{r}} {y_k}\nabla {h_k} < 0,k \in S}\\
		{{\tilde{r}} {y_k}\nabla {h_k} \ge 0,k \notin T}
		\end{array},{\tilde{r}} \in {\mathbb{R}^n}} \right.} \right\}
\end{equation}
Based on Eq.(\ref{eq:subdomain}), we redefine Eq.(\ref{eq:mattackr}) and Eq.(\ref{eq:attackt}) over the sub-domain $I_S$ of ${\bf{r}} $ as:
\begin{equation}\label{eq:submattackr}
{A_{h({{x}} ),{\tilde{r}} s}} = \sum\limits_{k \in S} {\frac{{ - {\tilde{r}} '\nabla {h_k}({{x}} )}}{{{h_k}({{x}} )}}}
\end{equation}

\begin{equation}\label{eq:subattackt}
\begin{array}{*{20}{l}}
{{\phi _s} = \mathop {\max }\limits_{{\tilde{r}}  \in {I_S}} {A_{h({{x}} ),{\tilde{r}} s}},}\\
{s.t.\,\,\,{{\left\| {\tilde{r}}  \right\|}_p} = 1}
\end{array}
\end{equation}

Now, we get ${{\phi _{h,{{x}} }} = \mathop {\max }\limits_{S \in P(S)} {\phi _s}}$. It's easy to know that:

\begin{equation}\label{eq:domain}
\begin{array}{*{20}{l}}
{{\phi _s} = \mathop {\max }\limits_{{\tilde{r}}  \in {I_S}} {A_{h({{x}} ),{\tilde{r}} s}},}\\
{s.t.\,\,\,{{\left\| {\tilde{r}}  \right\|}_p} = 1}
\end{array} \le \begin{array}{*{20}{l}}
{{\phi _s} = \mathop {\max }\limits_{{{e}}  \in {\mathbb{R}^n}} {A_{h({{x}} ),{{e}} s}},}\\
{s.t.\,\,\,{{\left\| {{e}}  \right\|}_p} = 1}
\end{array} = {\left\| {\sum\limits_{k \in S} {\frac{{{-}\nabla {h_k}({{x}} )}}{{{h_k}({{x}} )}}} } \right\|_q}
\end{equation}

The equality holds when the optimal ${{e}} ^*$ exactly locates in $I_S$. Now, if we want to prove that Eq.(\ref{eq:attackt}) $=$ Eq.(\ref{eq:comb}), we just need to prove that ${\phi _{{S^ * }}} = {\left\| {\sum\limits_{k \in {S^ * }} {\frac{{{-}\nabla {h_k}({{x}} )}}{{{h_k}({{x}} )}}} } \right\|_q}$, that is we need to prove that the optimal ${{e}} ^*$ for $S^*$ locates in ${I_{{S^ * }}}$. We can prove that by contradiction. That is we assume ${{e}} _{{S^ * }}^ *  \in {I_{S'}}(S' \ne {S^ * })$, then it is proved by Eq.(\ref{eq:contradicction}).

\begin{equation}\label{eq:contradicction}
\begin{array}{c}
{\left\| {\sum\limits_{k \in {S^*}} {\frac{{ - \nabla {h_k}({{x}} )}}{{{h_k}({{x}} )}}} } \right\|_q} = \sum\limits_{k \in {S^*}} {\frac{{ - {{e}} _{{S^ * }}^ * \nabla {h_k}({{x}} )}}{{{h_k}({{x}} )}}} \\
< \sum\limits_{k \in {S^*} \cap S'} {\frac{{ - {{e}} _{{S^ * }}^ * \nabla {h_k}({{x}} )}}{{{h_k}({{x}} )}}} \\
\le {\left\| {\sum\limits_{k \in {S^*} \cap S'} {\frac{{ - \nabla {h_k}({{x}} )}}{{{h_k}({{x}} )}}} } \right\|_q}\\
< {\left\| {\sum\limits_{k \in {S^*}} {\frac{{ - \nabla {h_k}({{x}} )}}{{{h_k}({{x}} )}}} } \right\|_q}
\end{array}
\end{equation}

\textbf{Proof of Eq.(\ref{eq:tbound}):}
We define the worst-case loss $l(h,{{z}}, \varepsilon)$ as:
\begin{equation}\label{eq:worstloss}
\begin{array}{l}
l({h},{{z}},\varepsilon ) = \mathop {\max }\limits_{{{z'}} \in N({{z}})} l({h},{{z'}}),\\
{\rm{where}}\,\,N({{z}}) = \left\{ {({{x'}},{{y'}})\left| {{{\left\| {{{x'}} - {{x}}} \right\|}_p} \le \varepsilon ,\,{{y'}} = {{y}}} \right.} \right\}.
\end{array}
\end{equation}
We first upperly bound $l(h,{{z}}, \varepsilon)$ defined in Eq.(\ref{eq:worstloss}) with the setting of linear classifier and hinge loss:
\begin{equation}
\begin{array}{c}
l(h,z,\varepsilon ) \le l(h,z) + \\
\mathop {\max }\limits_{{{\left\| {{r}} \right\|}_2} \le \varepsilon } {\left\| {\sum\limits_{k = 1}^m {{y_k}{{r}}' \cdot {{\bf{w}}_k}*\max \{ {\mathop{\rm sgn}} ({y_k}{{r}}' \cdot {{\bf{w}}_k}),0\} } } \right\|_2}\\
 \le l(h,z) + {C_{{\bf{W}},z}}\varepsilon. 
\end{array}
\end{equation}
The last step borrows the proof from Eq.\ref{eq:attackt} to Eq.\ref{eq:comb}. Then we have

\begin{equation}
\begin{array}{*{20}{l}}
{{R_\mathcal{D}}(A,\varepsilon ) - {R_{{Z^n}}}(A,\varepsilon )}\\
{ = \mathop {{\rm{ }}\mathbb{E}}\limits_{{Z^n} \leftarrow {\mathcal{D}^n},A} l(A({Z^n}),\mathcal{D},\varepsilon ) - \mathop {{\rm{ }}\mathbb{E}}\limits_{{Z^n} \leftarrow {\mathcal{D}^n},A} l(A({Z^n}),{Z^n},\varepsilon )}\\
{ = \mathop {{\rm{ }}\mathbb{E}}\limits_{\bar Z,E,A} \left[ {l(A({{\bar Z}_E}),{{\bar Z}_{\bar E}},\varepsilon ) - l(A({{\bar Z}_E}),{{\bar Z}_E},\varepsilon )} \right]\,,\,\,\,\,(\bar Z \leftarrow {\mathcal{D}^{n \times 2}})}\\
\begin{array}{l}
 = \mathop {{\rm{ }}\mathbb{E}}\limits_{\bar Z,E,A} \left[ {{f_{\bar Z}}(A({{\bar Z}_E}),E,\varepsilon )} \right]\\
  \,\,\,\,\,by\,\,LEMMA\,2\\
 \le \mathop {\inf }\limits_{t > 0} \frac{{I(A({{\bar Z}_E});E|\bar Z) + \mathop {{\rm{ }}\mathbb{E}}\limits_{\bar Z} \left[ {\log \mathop {{\rm{ }}\mathbb{E}}\limits_{{\bf{W}},E'} \left[ {{e^{t{f_{\bar Z}}({\bf{W}},E',\varepsilon )}}} \right]} \right]}}{t},\\
  \,\,\,\,by\,\,\,independence\\
 = \mathop {\inf }\limits_{t > 0} \frac{{CM{I_{\mathcal{D},A}} + \mathop {{\rm{ }}\mathbb{E}}\limits_{\bar Z} \left[ {\log \mathop {{\rm{ }}\mathbb{E}}\limits_{\bf{W}} \left[ {\prod\limits_{i = 1}^n {\mathop \mathbb{E}\limits_{{{E'}_i}} \left[ {{e^{\frac{t}{n}(l(W,{{({{\bar Z}_{E'}})}_i},\varepsilon ) - l({\bf{W}},{{({{\bar Z}_{\bar E'}})}_i},\varepsilon ))}}} \right]} } \right]} \right]}}{t},\\
 = \mathop {\inf }\limits_{t > 0} \frac{{CM{I_{\mathcal{D},A}} + \mathop {{\rm{ }}\mathbb{E}}\limits_{\bar Z} \left[ {\log \mathop {{\rm{ }}\mathbb{E}}\limits_{\bf{W}} \left[ {\prod\limits_{i = 1}^n {\mathop E\limits_{{{E'}_i}} \left[ {{e^{\frac{t}{n}(1 - 2{{E'}_i})(l({\bf{W}},{{\bar Z}_{i,1}},\varepsilon ) - l({\bf{W}},{{\bar Z}_{i,2}},\varepsilon ))}}} \right]} } \right]} \right]}}{t}
\end{array}\\
\,\,\,\,by\,\,LEMMA\,\,3\\
{ \le \mathop {\inf }\limits_{t > 0} \frac{{CM{I_{\mathcal{D},A}} + \mathop {{\rm{ }}\mathbb{E}}\limits_{\bar Z} \left[ {\log \mathop {{\rm{ }}\mathbb{E}}\limits_{\bf{W}} \left[ {\prod\limits_{i = 1}^n {{e^{\frac{{{t^2}}}{{2{n^2}}}{{(l({\bf{W}},{{\bar Z}_{i,1}},\varepsilon ) - l({\bf{W}},{{\bar Z}_{i,2}},\varepsilon ))}^2}}}} } \right]} \right]}}{t},}\\
{ \le \mathop {\inf }\limits_{t > 0} \frac{{CM{I_{\mathcal{D},A}}}}{t} + \frac{t}{{2n}}\mathop {{\rm{ }}\mathbb{E}}\limits_{\bar Z} \left[ {\mathop {\sup }\limits_{{\bf{W}} \in \mathcal{W}_A} \frac{1}{n}\sum\limits_{i = 1}^n {{{(l({\bf{W}},{{\bar Z}_{i,1}},\varepsilon ) - l({\bf{W}},{{\bar Z}_{i,2}},\varepsilon ))}^2}} } \right]}\\
{ \le \mathop {\inf }\limits_{t > 0} \frac{{CM{I_{\mathcal{D},A}}}}{t} + \frac{t}{{2n}}\mathop {{\rm{ }}\mathbb{E}}\limits_{Z \leftarrow \mathcal{D}} \left[ {\mathop {\sup }\limits_{{\bf{W}} \in \mathcal{W}_A} l{{({\bf{W}},Z,\varepsilon )}^2}} \right]}\\
{ \le \mathop {\inf }\limits_{t > 0} \frac{{CM{I_{\mathcal{D},A}}}}{t} + \frac{t}{{2n}}\mathop {{\rm{ }}\mathbb{E}}\limits_{Z \leftarrow \mathcal{D}} \left[ {\mathop {\sup }\limits_{{\bf{W}} \in \mathcal{W}_A} {{\left( {l({\bf{W}},Z) + {C_{{\bf{W}},Z}} \cdot \varepsilon } \right)}^2}} \right]}\\
{ = \sqrt {\frac{2}{n}CM{I_{\mathcal{D},A}} \cdot \mathop {{\rm{ }}\mathbb{E}}\limits_{Z \leftarrow \mathcal{D}} \left[ {\mathop {\sup }\limits_{{\bf{W}} \in \mathcal{W}_A} {{\left( {l({\bf{W}},Z) + {C_{{\bf{W}},Z}} \cdot \varepsilon } \right)}^2}} \right]} }
\end{array}
\end{equation}

\textbf{Proof of Eq.(\ref{eq:cmibound}):} Here we use $H$ to denote the entropy.
\begin{equation}
\begin{array}{l}
CM{I_{\mathcal{D},A}}\\
 = I(A;S,\bar Z) - I(A;\bar Z)\\
 = H(A) + H(S,\bar Z) - H(A,S,\bar Z) - H(A) - H(\bar Z) + H(A,\bar Z)\\
 = H(A,\bar Z) + H(S|\bar Z) - H(S) - H(A,\bar Z|S)\,\,\,\,:S\,\,is\,independent\,to\,Z\\
 = H(A,\bar Z) - H(A,\bar Z|S)\\
 \le H(A,\bar Z)\\
 \le H(A) + H(\bar Z)\\
 = H({\bf{W}}) + H(\bar Z)\\
 = ent({{\bf{w}}_1}, \cdots ,{{\bf{w}}_m}) + ent({\cal \mathcal{D}}_1, \cdots ,{\cal \mathcal{D}}_m) 
\end{array}
\end{equation}

\end{document}